\documentclass[10pt,twocolumn,twoside,journal]{IEEEtran} 
\usepackage{lipsum}
\usepackage{cite}
\usepackage{amsmath,amssymb,amsfonts}
\usepackage{algorithmic}
\usepackage{graphicx}
\usepackage{textcomp}
\usepackage{hyperref}
\usepackage{enumitem}
\hypersetup{hidelinks,colorlinks=true,allcolors=black,pdfstartview=Fit,breaklinks=true}
\usepackage{color}

\begin{document}
	
	\title{Distributed Control for a Multi-Agent System to Pass through a Connected Quadrangle Virtual Tube}
	
	\author{Yan Gao, Chenggang Bai, Quan Quan
	\thanks{Yan Gao, Chenggang Bai and Quan Quan are with the School of Automation Science and Electrical Engineering, Beihang University, Beijing 100191, P. R. China (email: buaa\_gaoyan@buaa.edu.cn; bcg@buaa.edu.cn; qq\_buaa@buaa.edu.cn)}}
	
	
\maketitle

\begin{abstract}
In order to guide the multi-agent system in a cluttered environment, a connected quadrangle virtual tube is designed for all agents to keep moving within it, whose basis is called the single trapezoid virtual tube. 
There is no obstacle inside the tube, namely the area inside the tube can be seen as a safety zone. 
Then, a distributed swarm controller is proposed for the single trapezoid virtual tube passing problem. This issue is resolved by a gradient vector field method with no local minima. Formal analyses and proofs are made to show that all agents are able to pass the single trapezoid virtual tube. Finally,  a modified controller is put forward for convenience in practical use. For the connected quadrangle virtual tube, a modified switching logic is proposed to avoid the deadlock and prevent agents from moving outside the virtual tube. Finally, the effectiveness of the proposed method is validated by numerical simulations and real experiments.
\end{abstract}


\section{Introduction}
Recently, it is a typical challenge for a multi-agent system to pass through a cluttered environment and reach the appointed area \cite{Chung(2018)}. The multi-agent system, especially the multi-multicopter system, must be capable of planning movements for all agents reliably and safely. Not only should each agent avoid collision with obstacles, but the agents also need to avoid collision with each other \cite{Quan(2017)}.

Numerous approaches in the existing literature have been put forward for controlling a multi-agent system to operate in a cluttered environment. For a large multi-agent system, it can be classified into two types according to the number of agents \cite{Chung(2018)}: \emph{formation} less than one hundred and \emph{swarm} up to thousands. The difference in quantity leads to differences in behavior and control methods. The \emph{formation} consists of cooperative interactions among all agents, the relationship of which is well-defined for designated objectives \cite{Oh(2015)}. Each agent in the formation usually remains a prespecified pose and makes the formation stable and robust \cite{Khan(2016)}, \cite{Zhao(2019)},\cite{chen2021robust}.
The affine formation maneuver control is especially suitable for the transformation control \cite{Xu(2020)}. However, the formation is not perfect in all circumstances. If there are hundreds or even thousands of agents in the multi-agent system, the formation will be too big to maintain feasibility. The increase in quantity leads to the expansion of physical size, which is infeasible in many narrow spaces. Besides, when some agents need to change their locations, it may cause chaos in the formation, and the formation controller will become too complex to maintain the formation stability. 
Under this circumstance, a swarm is the best choice, which generally displays emergent behavior arising from local interactions among the agents \cite{Chung(2018)}. The methods designed for the swarm also suit the formation, which implies that the swarm has a wider application range and suits more situations.

\begin{figure}[!t]
	\centerline{\includegraphics[width=\columnwidth]{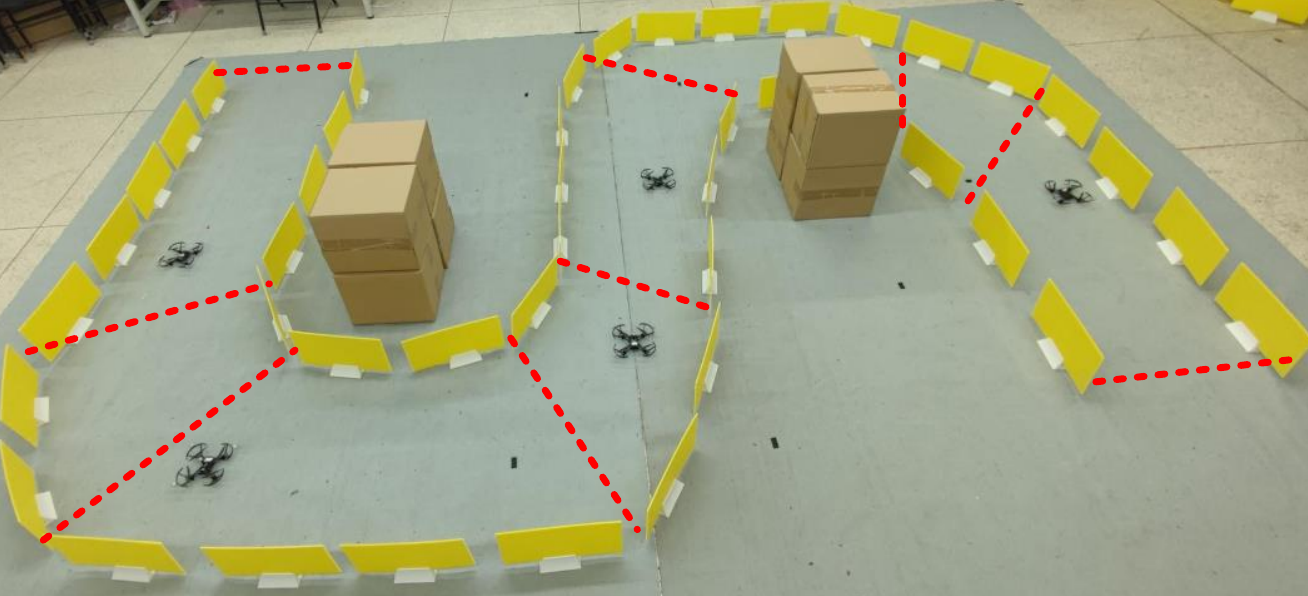}}
	\caption{A connected quadrangle virtual tube is designed for guiding multiple multicopters in a cluttered environment.}
	\label{First}
\end{figure}


For \emph{swarm} navigation and control, the control-based methods are widely used because of their simplicity and accessibility \cite{Wolf(2008)}.  Although the control-based methods possess a weaker control performance compared with the multi-agent trajectory planning  \cite{Zhou(2020)}, \cite{Park(2020)}, they are more suitable for the large-scale multi-agent system.
Control-based methods directly guide the agents' movement according to the global path and current local information \cite{Quan(2021)}, \cite{Miao(2016)}, \cite{Wang(2017)}. Distributed multi-agent trajectory planning needs any of the agents to share its planned trajectory with others via wireless communication, which brings a huge communication pressure when the number of agents increases \cite{Zhou(2020)}.  Control-based methods usually use a simple controller formula to react to obstacles or other agents, which have a good quality to achieve a fast and reactive response to a dynamic environment and a low demand for computation and communication resources. In \cite{santilli2022multirobot}, the authors solve the problem of coordinating the motion of a team of robots with limited field of view in the traditional gradient systems. The conclusions in this paper is helpful for our future work. Besides, the control barrier function (CBF) method is also popular in recent years, which is summarized as a quadratic programming (QP) problem with better performance and higher demand on the computational resources \cite{Wang(2017)}. 

The method proposed in this paper is a type of artificial potential field (APF) method belonging to control-based methods \cite{Khatib(1985)}. The APF method 
can be seen as a gradient vector field method. In contrast, there are also non-potential vector field methods, whose curls are non-zero. However, the function forms of the non-potential vector field methods are limited, the stability proof of which are also non-trivial \cite{Panagou(2014)}, \cite{Panagou(2016)}. Compared with the CBF method, the APF method is especially suitable for dealing with multi-objective compositions at the same time,
which is rather complicated for the CBF method, as multiple hard safety constraints may cause no feasible solution. Existing literature is limited to the combination of similar and complementary objectives, such as collision avoidance and connectivity maintenance \cite{Wang(2016)}. For the APF method, each control objective can be described as a potential function.
By summing up all potential functions, the corresponding vector field is directly generated with a negative gradient operation. Nevertheless, inappropriate definitions of the potential field will cause various problems, in which the most serious is local minima \cite{Hernandez(2011)}. The local minima problem is the appearance of unexpected equilibrium points where the composite potential field vanishes. 

Motivated by the current studies, we present a \textit{connected quadrangle virtual tube} to guide the multi-agent system in a cluttered environment, whose basis is a single trapezoid virtual tube. The term ``virtual tube" appears in the AIRBUS’s Skyways project
\cite{AIRBUS}. In our previous work \cite{Quan(2021)}, the \textit{straight-line virtual tube} is proposed for the air traffic control as flight routes are usually composed of several line segments. There is no obstacle inside the virtual tube, which implies that the area inside can be seen as a safety zone. 
In this paper, the concept of the virtual tube is generalized. The connected quadrangle virtual tube is more suitable for guiding the multi-agent system to move within a narrow corridor, through a window or a doorframe. Besides, as a trapezoid or a quadrangle becomes a rectangle when edges are perpendicular to each other, this novel virtual tube can also be used for the air traffic control.  As shown in Fig. \ref{First}, the concept of the connected quadrangle virtual tube is similar to the lane for autonomous road vehicles in \cite{Rasekhipour(2016)} and the corridor for a multi-UAV system in \cite{Tony(2020)}, \cite{Liu(2017)}.

For the connected quadrangle virtual tube, two problems are summarized, namely \textit{connected quadrangle virtual tube planning problem} and \textit{connected quadrangle virtual tube passing problem}. This paper only aims to solve the latter one. For the former problem, the virtual tube can be automatically generated from a given environment with the traditional path planning algorithm \cite{Likhachev(2004)}, \cite{Kavraki(1996)}. Another feasible approach is to expand an existing path, which performs like a ``teach-and-repeat" system \cite{Gao(2019)}. When there are $M$ agents, the distributed multi-agent trajectory planning has to plan $M$ trajectories, while our method only needs one trajectory to generate a virtual tube. The connected quadrangle virtual tube passing problem is solved in this paper with a distributed vector field method, which can be seen as a modified APF method. 

From the straight-line virtual tube in \cite{Quan(2021)} to the connected quadrangle virtual tube, the main challenge is the controller design. As the width of the virtual tube in this paper is not immutable, the controller for a single straight-line virtual tube cannot directly apply to a single trapezoid virtual tube. Otherwise, there may exist a deadlock problem. Besides, the switching logic between adjacent quadrangles should be designed carefully to avoid the deadlock.
There is no uncertainty considered in this paper, and all agents are able to obtain the information clearly and execute the velocity command exactly. In real practice, the \emph{separation theorem} in our previous work \cite{Fu(2021)} can be introduced to deal with the position estimate noise, the broadcast delay, the packet loss and the transient performance caused by some filters and observers. In short, all uncertainties are considered in the design of the safety radius, and the controller is irrelevant to uncertainties. Besides, in our recent work \cite{Gao2022}, it is shown that the cohesion behavior and the velocity alignment behavior of the flocking algorithm are able to reduce the influence of the \emph{position measurement drift} and the \emph{velocity measurement error}, respectively. Relative control terms can be added to the controller proposed in this paper.

In this paper, two models for agents and trapezoid virtual tubes are first proposed. Then, two problems to be solved
are defined. 
A new type of Lyapunov function, called \emph{Line Integral Lyapunov Function}, is designed to guide agents to reach the finishing line. Besides, the single panel method and a Lyapunov-like barrier function are proposed for restricting agents to moving inside the virtual tube and avoiding collision with each other. Finally, a distributed swarm controller with a necessary saturation constraint is designed. For practical use, a modified swarm controller with a similar control effect is also presented. For the connected quadrangle virtual tube passing problem, a modified switching logic is proposed. We prove that the multi-agent system is able to pass through the single trapezoid virtual tube and the connected quadrangle virtual tube based on the \emph{invariant set theorem} \cite[p. 69]{Slotine(1991)}. The major contributions of this paper are summarized as follows: 

\begin{itemize}[leftmargin=*]
	\item Based on the straight-line virtual tube introduced in our previous work [R1], the connected quadrangle virtual tube is proposed, which is especially suitable for guiding a multi-agent system in cluttered environments, such as moving within a narrow corridor, passing through a window or a doorframe. The connected quadrangle virtual tube makes a significant advance over existing planning and formation methods. Also, this work opens a new way of planning from a single agent having a one-dimensional path to multiple agents sharing a two-dimensional virtual tube. 
	\item A local minima-free potential field controller is proposed for guiding the agents inside the trapezoid virtual tube. As the width of the virtual tube in this paper is not immutable, the proposed controller is different from the controller in [R1]. Otherwise, there may exist a deadlock problem. Besides, we present a switching logic to transfer the quadrangle control problem to several single trapezoid control problems. When agents are transitioning between trapezoids, the switching logic is designed properly to avoid the deadlock. 
	\item A formal proof is proposed to show that there is no collision among agents, and all agents can keep within the virtual tube and pass through the finishing line without getting stuck. The key to the proof is the use of the single panel method, which is a part of the final local minima-free potential field function. The single panel method ensures that the angle between the orientation of the virtual tube keeping term and the orientation of the line approaching term is always smaller than $90^{\circ}$.
\end{itemize}

\section{Problem Formulation}

\subsection{Agent Model}

In this paper, the multi-agent system considered consists of $M$  agents in a horizontal plane $\mathbb{R}^2$. Each agent is velocity-controlled with a single integral holonomic kinematics
\begin{equation}
	\dot{\mathbf{{p}}}_{i} =\mathbf{v}_{\text{c},i}, \label{SingleIntegral}
\end{equation}
in which $\mathbf{v}_{\text{c},i}\in {{\mathbb{R}}^{2}}$, $\mathbf{p}_{i}\in {{\mathbb{R}}^{2}}$ are the velocity command and position of the $i$th agent, respectively. Besides, $v_{\text{m},i}>0$ is set as the maximum permitted speed of the $i$th agent. Hence it is necessary to make $\mathbf{v}_{\text{c},i}$ subject to a saturation function
$
	\mathbf{v}_{\text{c},i}=\text{sa}{\text{t}}\left(\mathbf{v}^{\prime}_{\text{c},i},{v_{\text{m},i}}\right) ={{\kappa}_{{v_{\text{m},i}}}}\left(\mathbf{v}^{\prime}_{\text{c},i}\right)\mathbf{v}^{\prime}_{\text{c},i},
$
where $\mathbf{v}^{\prime}_{\text{c},i}\in {{\mathbb{R}}^{2}}$ is the original velocity command and
\begin{equation*}
	\text{sa}{\text{t}}\left( \mathbf{v}^{\prime}_{\text{c},i},{v_{\text{m},i}}\right) \triangleq
	\begin{cases}
		\mathbf{v}^{\prime}_{\text{c},i}  & 	\left \Vert \mathbf{v}^{\prime}_{\text{c},i}\right \Vert \leq {v_{\text{m},i}}\\ 
		{v_{\text{m},i}}\frac{\mathbf{v}^{\prime}_{\text{c},i}}{\left \Vert \mathbf{v}^{\prime}_{\text{c},i}\right \Vert } & \left \Vert \mathbf{v}^{\prime}_{\text{c},i}\right \Vert  >{v_{\text{m},i}}
	\end{cases},
\end{equation*}
\begin{equation*}
	{{\kappa }_{{v_{\text{m},i}}}}\left( \mathbf{v}^{\prime}_{\text{c},i}\right) \triangleq
	\begin{cases}
		1 & \left \Vert \mathbf{v}^{\prime}_{\text{c},i}\right \Vert \leq {v_{\text{m},i}}\\ 
		\frac{{v_{\text{m},i}}}{\left \Vert \mathbf{v}^{\prime}_{\text{c},i}\right \Vert } & 	\left \Vert \mathbf{v}^{\prime}_{\text{c},i}\right \Vert >{v_{\text{m},i}}
	\end{cases}.
\end{equation*}
It is obvious that $0<{{\kappa }_{{v_{\text{m},i}}}}\left( \mathbf{v}^{\prime}_{\text{c},i}\right)
\leq 1$. In the following, ${{\kappa }_{{v_{\text{m},i}}}}\left(\mathbf{v}^{\prime}_{\text{c},i}\right) $
will be written as ${{\kappa }_{{v_{\text{m},i}}}}$ for short. Besides, $\mathbf{v}^{\prime}_{\text{c},i}$ and  $\mathbf{v}_{\text{c},i}$ always have the same direction.
When an agent is modeled as a single integrator just like (\ref{SingleIntegral}), such as multicopters, helicopters and certain types of wheeled robots equipped with omni-directional wheels \cite{Quan(2017)}, the designed velocity command $\mathbf{v}_{\text{c},i}$ can be directly applied to control the agent. When the model considered is more complicated, such as a second-order integrator model, additional control laws are necessary \cite{Rezende(2020)}. Besides, in our previous work \cite{Quan(2021)}, we propose a \emph{filtered position model} converting a second-order model to a first-order model just like \eqref{SingleIntegral}.

\subsection{Connected Quadrangle Virtual Tube Model}

As any quadrangle can be considered to be contained in a trapezoid, we first propose a model for the single trapezoid virtual tube. Then the model for the connected quadrangle virtual tube is presented.

\begin{figure}[tbp]
	\centering
	\includegraphics[scale=1]{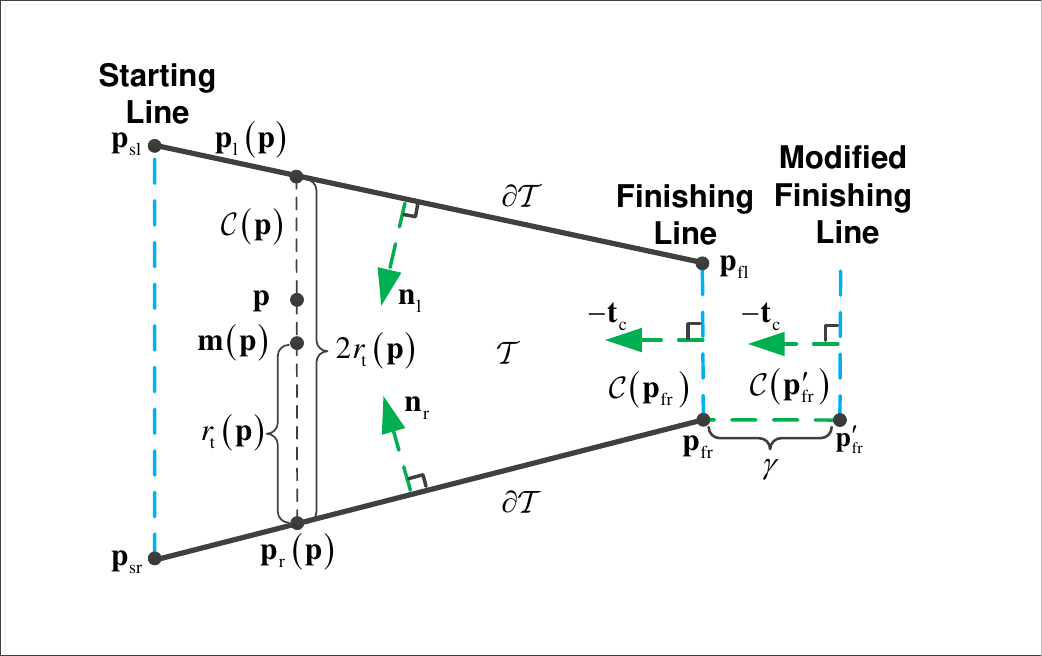}
	\caption{Single trapezoid virtual tube.}
	\label{airwaytube}
\end{figure}

\begin{itemize}[leftmargin=*]
	\item \textbf{Single Trapezoid Virtual Tube}. As shown in
	Fig. \ref{airwaytube}, a single trapezoid virtual tube $\mathcal{T}\left( {{\mathbf{p}}_{\text{fr}},{\mathbf{p}}_{\text{fl
		}},{\mathbf{p}}_{\text{sl}},{\mathbf{p}}_{\text{sr}}}\right) $ locates in a horizontal plane ${\mathbb{R}}^{2}$. Points ${{\mathbf{p}}_{\text{fl}},{\mathbf{p}}_{\text{fr}%
		},{\mathbf{p}}_{\text{sl}},{\mathbf{p}}_{\text{sr}}}\in {{\mathbb{R}}^{2}}$ are four vertices of the trapezoid. Sometimes, $\mathcal{T}\left( {{\mathbf{p}}_{\text{fr}},{\mathbf{p}}_{\text{fl
		}},{\mathbf{p}}_{\text{sl}},{\mathbf{p}}_{\text{sr}}}\right) $
	will be written as $\mathcal{T}$ for short. There are two parallel bases in the trapezoid, namely $\left[ {{\mathbf{p}}_{\text{fr}},{\mathbf{p}}_{\text{fl}}}\right] $ and $\left[ {{\mathbf{p}}_{\text{sr}},{%
			\mathbf{p}}_{\text{sl}}}\right] $. Line segments $\left[ {{\mathbf{p}}%
		_{\text{fr}},{\mathbf{p}}_{\text{sr}}}\right] $ and $\left[ {{\mathbf{p}}_{%
			\text{fl}},{\mathbf{p}}_{\text{sl}}}\right] $ are two legs. 
	Then the trapezoid virtual tube $\mathcal{T}$ is expressed as
	\begin{align*}
		\mathcal{T}=&\left \{ \mathbf{x}\in {{\mathbb{R}}^{2}} :
		\mathbf{n}_{\text{l}}^{\text{T}}\left( \mathbf{x}-{{\mathbf{p}}_{\text{fl}}}%
		\right) \geq 0,\mathbf{n}_{\text{r}}^{\text{T}}\left( \mathbf{x}-{{\mathbf{p}%
			}_{\text{fr}}}\right) \geq 0,\right.\\ 
		&\left.-\mathbf{t}_{\text{c}}^{\text{T}}\left( \mathbf{%
			x}-{{\mathbf{p}}_{\text{fr}}}\right) \geq 0,-\mathbf{t}_{\text{c}}^{\text{T}%
		}\left( \mathbf{x}-{{\mathbf{p}}_{\text{sr}}}\right) \leq 0\right \}.
	\end{align*}
	And the boundary of $\mathcal{T}$ is shown as
	\begin{align*}
		\partial \mathcal{T}=\left \{ \mathbf{x}\in \mathcal{T}: 
		\mathbf{n}_{\text{l}}^{\text{T}}\left( \mathbf{x}-{{\mathbf{p}}_{\text{fl}}}%
		\right) =0\cup \mathbf{n}_{\text{r}}^{\text{T}}\left( \mathbf{x}-{{\mathbf{p}%
			}_{\text{fr}}}\right) =0\right \},
	\end{align*}
	where unit vectors $\mathbf{n}_{\text{l}},\mathbf{n}_{\text{r}}\in {{\mathbb{%
				R}}^{2}}$ are linearly independent of unit vector $\mathbf{t}_{\text{c}%
	}\in {{\mathbb{R}}^{2}}$. Moreover it is obtained that 
	$\mathbf{n}_{\text{l}}^{\text{T}}\left( {{\mathbf{p}}_{\text{sl}}}-{{\mathbf{p}}_{\text{fl}}}\right) =0,$ $\mathbf{n}_{\text{r}}^{\text{T}}\left( {{\mathbf{p}}_{\text{sr}}}-{{\mathbf{p}}_{\text{fr}}}\right) =0 ,$ $\mathbf{t}_{\text{c}}^{\text{T}}\left( {{\mathbf{p}}_{\text{sr}}-{\mathbf{p}}_{\text{sl}}}\right) =0.
	$	
	\item \textbf{Cross Section}. For any point $\mathbf{p}\in \mathcal{T},$ a \emph{%
		cross section }passing $\mathbf{p\ }$is defined as
	\begin{align*}
		\mathcal{C}\left( \mathbf{p}\right) =\left \{ \mathbf{x}\in {{\mathbb{%
					R}}^{2}}: \mathbf{t}_{\text{c}}^{\text{T}}\left( \mathbf{x}-{{%
				\mathbf{p}}}\right) =0\cap \mathcal{T}\right \} .
	\end{align*}
	Here, $\mathcal{C}\left( {{\mathbf{p}}_{\text{f}}}\right) =\mathcal{C}%
	\left( {{\mathbf{p}}_{\text{fr}}}\right) =\mathcal{C}\left( {{\mathbf{p}}_{%
			\text{fl}}}\right) $ is called the \emph{finishing line} or \emph{finishing
		cross section}, where $\mathbf{p}_{\text{f}}$ is any point located on the finishing line. 
	Furthermore, points $\mathbf{p}_{\text{l}}\left( \mathbf{p}\right),\mathbf{p}_{\text{r}}\left( \mathbf{p}\right)\in {\mathbb{R}}^{2}$ are the
	intersection points of the cross section $ \mathcal{C}\left( \mathbf{p}\right)$ with the tube boundary $\partial \mathcal{T}$, namely
$
		\mathbf{p}_{\text{l}}\left( \mathbf{p}\right)  =\left \{  \mathbf{x}%
		\in \mathcal{C}\left( \mathbf{p}\right):\mathbf{n}_{\text{l}}^{%
			\text{T}}\left( \mathbf{x}-{{\mathbf{p}}_{\text{fl}}}\right) =0\right \} $,
		$
		\mathbf{p}_{\text{r}}\left( \mathbf{p}\right)  =\left \{  \mathbf{x}
		\in \mathcal{C}\left( \mathbf{p}\right) : \mathbf{n}_{\text{r}}^{
			\text{T}}\left( \mathbf{x}-{{\mathbf{p}}_{\text{fr}}}\right) =0\right \} .
$
	The width of $\mathcal{C}\left( \mathbf{p}\right)$ is denoted by 2$r_{\text{t}}\left( \mathbf{p}%
	\right) ,$ which is defined as
	\begin{align*}
		r_{\text{t}}\left( \mathbf{p}\right) \triangleq \frac{1}{2}\left \Vert 
		\mathbf{p}_{\text{r}}\left( \mathbf{p}\right) -\mathbf{p}_{\text{l}}\left( 
		\mathbf{p}\right) \right \Vert .
	\end{align*}
	The width of the trapezoid virtual tube is further defined as $r_{\text{t}}\left( \mathcal{T}\right) =\underset{\mathbf{p}\in \mathcal{T}}{\inf }r_{\text{t}}\left( \mathbf{p}\right)$.
	Besides, the middle point $\mathbf{m}\left( \mathbf{p}\right) \in \mathcal{C}\left( \mathbf{p}\right) $ of the cross section $\mathcal{C}\left( \mathbf{p}\right)$ is defined as
$
		\mathbf{m}\left( \mathbf{p}\right) \triangleq \frac{1}{2}\left( \mathbf{p}_{\text{l}}\left( \mathbf{p}\right)
		+\mathbf{p}_{\text{r}}\left( \mathbf{p}\right) \right) .
$
	

	\begin{figure}[tbp]
		\centering \includegraphics[scale=1]{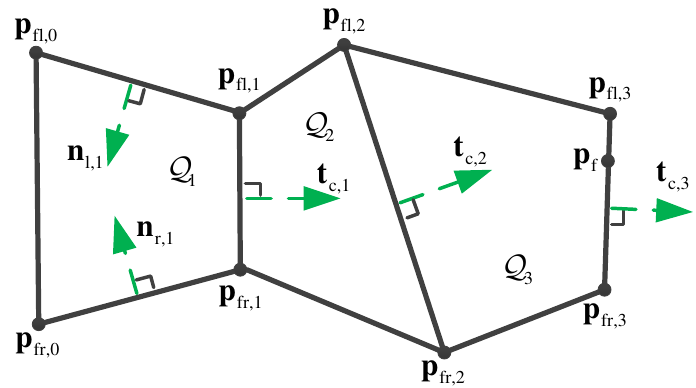}
		\caption{Connected quadrangle virtual tube.}
		\label{ConnectedQuadrangle_nomark}
	\end{figure}
	
	\item \textbf{Connected Quadrangle Virtual Tube}. As shown in Fig. \ref{ConnectedQuadrangle_nomark}, we further define the model of a connected quadrangle virtual tube with the definition of a single trapezoid virtual tube. Here a connected quadrangle virtual tube $\mathcal{Q}$ composed of $N$ quadrangles is proposed as 
	\begin{equation}
		\mathcal{Q}=\underset{q=1,\cdots ,N}\cup \mathcal{Q}%
		_{q}\left( {{\mathbf{p}}_{\text{fr,}q},{\mathbf{p}}_{\text{fl,}q},{\mathbf{p}%
			}_{\text{fl,}q-1},{\mathbf{p}}_{\text{fr,}q-1}}\right), \label{ConnectedQuadrangle}
	\end{equation}
	where $\mathcal{Q}_{q}$ is the $q$th quadrangle, $q=1,\cdots ,N$. And the boundary of $\mathcal{Q}$ is
	\begin{equation*}
		\partial \mathcal{Q}= \underset{q=1,\cdots ,N}\cup
		\partial \mathcal{Q}_{q}\left( {{\mathbf{p}}_{\text{fr,}q},{\mathbf{p}}_{\text{fl,}q},{\mathbf{p}%
			}_{\text{fl,}q-1},{\mathbf{p}}_{\text{fr,}q-1}}\right),
	\end{equation*}
	where $\partial\mathcal{Q}_{q}$ is the boundary of the $q$th quadrangle.
	The quadrangle $\mathcal{Q}_{q}$ has two similar but different definitions in the following
	\begin{align*}
		\mathcal{Q}_{q}^{\prime }=&\left \{\mathbf{x}\in {{\mathbb{R}}^{2}}%
		: \mathbf{n}_{\text{l,}q}^{\text{T}}\left( \mathbf{x}-{{\mathbf{p}%
			}_{\text{fl,}q}}\right) \geq 0,\mathbf{n}_{\text{r,}q}^{\text{T}}\left( 
		\mathbf{x}-{{\mathbf{p}}_{\text{fr,}q}}\right) \geq 0,\right.\\
		&\left.-\mathbf{t}_{\text{c,}%
			q}^{\text{T}}\left( \mathbf{x}-{{\mathbf{p}}_{\text{fr,}q}}\right) \geq 0,%
		\mathbf{t}_{\text{c,}q-1}^{\text{T}}\left( \mathbf{x}-{{\mathbf{p}}_{\text{%
					fl,}q-1}}\right) \geq 0\right \}
	\end{align*}%
	or 
	\begin{align*}
		\mathcal{Q}_{q}^{\prime \prime }=&\left \{\mathbf{x}\in {{\mathbb{R}}
			^{2}}: \mathbf{n}_{\text{l,}q}^{\text{T}}\left( \mathbf{x}-{{%
				\mathbf{p}}_{\text{fl,}q}}\right) \geq 0,\mathbf{n}_{\text{r,}q}^{\text{T}%
		}\left( \mathbf{x}-{{\mathbf{p}}_{\text{fr,}q}}\right) \geq 0,\right.\\
		&\left.-\mathbf{t}_{%
			\text{c,}q}^{\text{T}}\left( \mathbf{x}-{{\mathbf{p}}_{\text{fr,}q}}\right)
		\geq 0,\mathbf{t}_{\text{c,}q-1}^{\text{T}}\left( \mathbf{x}-{{\mathbf{p}}_{%
				\text{fr,}q-1}}\right) \geq 0\right \} .
	\end{align*}
	And $\partial \mathcal{Q}_{q}$ is shown as 
	\begin{align*}
		\partial \mathcal{Q}_{q}\!=\!\left \{ \mathbf{x}\in \mathcal{Q}_{q}\!: \!
		\mathbf{n}_{\text{l,}q}^{\text{T}}\left( \mathbf{x}-{{\mathbf{p}}_{\text{fl,}q}}\right) = 0\!\cup\! \mathbf{n}_{\text{r,}q}^{\text{T}}\left( \mathbf{x}-{{\mathbf{p}}_{\text{fr,}q}}\right) = 0\right \}.
	\end{align*}
	The unit vectors $\mathbf{n}_{\text{l,}q},
	\mathbf{n}_{\text{r,}q}\in {{\mathbb{R}}^{2}}$ are linearly independent of unit vector $\mathbf{t}_{\text{c,}q}\in {{\mathbb{R}}^{2}}$, $q=1,\cdots ,N$. Different from the trapezoid, two bases in the quadrangle, $\left[ {{\mathbf{p}}_{\text{fr},q},{\mathbf{p}}_{\text{fl},q}}\right] $ and $\left[ {{\mathbf{p}}_{\text{sr},q},{\mathbf{p}}_{\text{sl},q}}\right] $, are not necessarily parallel. Without loss of generality, as any quadrangle can be contained in a corresponding trapezoid, we let the quadrangle $\mathcal{Q}_{1}$ be a trapezoid, namely
	\begin{equation}
		\mathcal{Q}_{1}\left( {{%
				\mathbf{p}}_{\text{fr,}1},{\mathbf{p}}_{\text{fl,}1},{\mathbf{p}}_{\text{fl,}%
				0},{\mathbf{p}}_{\text{fr,}0}}\right) =\mathcal{T}\left( {{\mathbf{p}}_{%
				\text{fr,}1},{\mathbf{p}}_{\text{fl,}1},{\mathbf{p}}_{\text{fl,}0},{\mathbf{p%
			}}_{\text{fr,}0}}\right) . \label{trapezoidquadrangle}
	\end{equation}
	Obviously, it is obtained that $\mathcal{Q}_{q}\cap \mathcal{Q}_{q+1} =\left[ {{\mathbf{p}}_{\text{fl,}q\text{,}},{\mathbf{p}}_{\text{fr,}q}}\right] ,\mathcal{Q}_{p}\cap \mathcal{Q}_{r} =\varnothing ,\left \vert p-r\right
	\vert \geq 2$,
	where $q=1,\cdots ,N-1,\ p=1,\cdots ,N,\ r=1,\cdots ,N.$ 
\end{itemize}

\subsection{Two Areas around an Agent}

Similarly to our previous work \cite{Quan(2021)}, two types of circular areas around an agent are introduced for the avoidance control, namely safety area and avoidance area. 
At the time $t>0$, the \emph{safety area} $\mathcal{S}_{i}$ of the $i$th agent is defined as
$
	\mathcal{S}_{i}\left(t\right)=\left \{ \mathbf{x}\in {{\mathbb{R}}^{2}}: \left \Vert \mathbf{x}-\mathbf{p}_{i}\left(t\right)\right \Vert \leq r_{\text{s}} \right \} ,
$
where $r_{\text{s}}>0$ is the \emph{safety radius}. For all agents, no \emph{conflict} with each other implies that $\mathcal{S}_{i}\cap \mathcal{S}_{j}=\varnothing$,
namely $\left \Vert \mathbf{p}_{i}-\mathbf{p}_{j}\right \Vert >2r_{\text{s}}$,
where $i,j=1,\cdots ,M,i\neq j$. Besides, the \emph{avoidance area} is defined for starting the avoidance control. At the time $t>0$, the avoidance area $\mathcal{A}_{i}$ of the $i$th agent is defined as 
$
	\mathcal{A}_{i}\left(t\right)=\left \{ \mathbf{x}\in {{\mathbb{R}}^{2}}:\left
	\Vert \mathbf{x}- \mathbf{p}_{i}\left(t\right)\right \Vert \leq r_{\text{a}}
	\right \}  ,
$
where $r_{\text{a}}>0$ is the \emph{avoidance radius}. For collision avoidance with any pair of agents, if there exist $\mathcal{A}_{i}\cap \mathcal{S}_{j}\neq \varnothing$ and $\mathcal{A}_{j}\cap \mathcal{S}_{i}\neq \varnothing$,
namely $\left \Vert \mathbf{p}_{i}-\mathbf{p}_{j}\right \Vert \leq r_{\text{a}}+r_{\text{s}}$,
then the $i$th and $j$th agents should avoid each other. The set $\mathcal{N}_{\text{m},i}$ is defined as the collection of all labels of other agents whose safety areas have intersection with the avoidance area of the $i$th agent, namely 
$
	\mathcal{N}_{\text{m},i}=\left \{ j: \mathcal{A}_{i} \cap \mathcal{S}_{j}
	\neq \varnothing\right \},
$
where $i,j=1,\cdots ,M,i\neq j$. Besides, when the $j$th agent or the tube boundary just enters the avoidance area $\mathcal{A}_{i}$ of the $i$th agent, it is required that there is no conflict in the beginning. Therefore, at least we set that 
$r_{\text{a}}>r_{\text{s}}$. 

\subsection{Virtual Tube Passing Problem Formulation}

With the descriptions above, some extra assumptions are proposed to get the main problem of this paper.

\textbf{Assumption 1}. The agents' initial conditions satisfy $\mathcal{S}_{i}\left( 0\right) \subset \mathcal{T}$ (for the single trapezoid virtual tube
passing problem) or $\mathcal{S}_{i}\left( 0\right) \subset\mathcal{Q}$ (for the connected quadrangle virtual tube passing problem), and $\mathcal{S}_{i}\left( 0\right)\cap \mathcal{S}_{j}\left( 0\right)=\varnothing$, where $i,j=1,\cdots ,M, i\neq j$.

\textbf{Assumption 2}. Once an agent arrives at the finishing line $\mathcal{C}%
\left( {{\mathbf{p}}_{\text{f}}}\right) =\left[ {{\mathbf{p}}_{\text{fr,}},{%
		\mathbf{p}}_{\text{fl}}}\right] $ (for the single trapezoid virtual tube passing problem, ${{\mathbf{p}}_{\text{f}}}\in 
\left[ {{\mathbf{p}}_{\text{fr}},{\mathbf{p}}_{\text{fl}}}\right] $) or $\mathcal{C}%
\left( {{\mathbf{p}}_{\text{f}}}\right)=
\left[ {{\mathbf{p}}_{\text{fr,}N},{\mathbf{p}}_{\text{fl,}N}}\right] $ (for the
connected quadrangle virtual tube passing problem, ${{\mathbf{p}}_{\text{f}}}\in 
\left[ {{\mathbf{p}}_{\text{fr,}N},{\mathbf{p}}_{\text{fl,}N}}\right] $),
it will quit the virtual tube not to affect other agents behind. Mathematically,
given ${\epsilon }_{\text{0}}>0,$ an agent arrives near $\mathcal{C}\left( {{\mathbf{p}}_{\text{f}}}\right)$ if 
\begin{equation}
	-\mathbf{t}_{\text{c}}^{\text{T}}\left( \mathbf{p}_{i}-{{\mathbf{p}}_{\text{f%
	}}}\right) \leq {\epsilon }_{\text{0}},  \label{arrivialairway}
\end{equation}
where $-\mathbf{t}_{\text{c}}$ is the moving direction of the single trapezoid virtual tube or the last quadrangle virtual tube.

Based on \textit{Assumptions 1, 2}, two main problems are stated in the following.

\begin{itemize}[leftmargin=*]
	\item \textbf{Single trapezoid virtual tube passing problem}. Under \textit{Assumptions 1, 2}, design the velocity command $\mathbf{v}_{\text{c},i}$ to guide all agents to pass the finishing line $\mathcal{C}\left( {{%
			\mathbf{p}}_{\text{f}}}\right) =\left[ {{\mathbf{p}}_{\text{fr}
		},{\mathbf{p}}_{\text{fl}}}\right]$ of the trapezoid virtual tube $\mathcal{T}$, meanwhile
	avoiding collision with other agents ($\mathcal{S}_{i}\left(t\right)\cap  \mathcal{S}_{j}\left(t\right)=\varnothing $) and keeping within the virtual tube ($\mathcal{S}_{i}\left(t\right)\cap \partial\mathcal{T}=\varnothing$), where  $i,j=1,\cdots ,M,i \neq j,t>0$.
	
	\item \textbf{Connected quadrangle virtual tube passing problem}. Under 
	\textit{Assumptions 1, 2}, design the velocity command $\mathbf{v}_{\text{c},i}$ to guide all agents to pass the finishing line $\mathcal{C}\left( {{%
			\mathbf{p}}_{\text{f}}}\right) =\left[ {{\mathbf{p}}_{\text{fr,}
			N},{\mathbf{p}}_{\text{fl,}N}}\right] $ of the connected quadrangle virtual tube $\mathcal{Q}$,
	meanwhile avoiding collision with other agents  ($\mathcal{S}_{i}\left(t\right)\cap  \mathcal{S}_{j}\left(t\right)=\varnothing $)  and keeping within the virtual
	tube ($\mathcal{S}_{i}\left(t\right)\cap \partial\mathcal{Q}=\varnothing$), 
	where  $i,j=1,\cdots ,M,i \neq j,t>0$.
\end{itemize}

\textbf{Remark 1}. Since there exists the relationship (\ref{trapezoidquadrangle}), the single trapezoid virtual tube passing problem can be considered as the first quadrangle virtual tube passing problem of the connected quadrangle virtual tube passing problem. Besides, the number of agents is not limited, as long as the virtual tube can contain these agents in the beginning.

\section{Distributed Control for Passing a Single Trapezoid Virtual Tube}
\subsection{Preliminaries}

\subsubsection{Line Integral Lyapunov Function for Vectors}
A new type of Lyapunov function for vectors, called \emph{Line Integral Lyapunov Function}, is designed as 
\begin{equation}
	V_{\text{li}}\left(\mathbf{y}\right)=\int_{S_{\mathbf{y}}}\text{sa}{\text{t}}%
	\left(\mathbf{x},a\right)^{\text{T}}\text{d}\mathbf{x},  \label{Vli0}
\end{equation}
where $a>0,$ $\mathbf{x}\in \mathbf{\mathbb{R}}^{2},$ $S_{\mathbf{y}}$ is a smooth curve from $\mathbf{0}$ to $\mathbf{y} \in \mathbb{R}^{2}$. In the following lemma, we will show its properties.

\textbf{Lemma 1} \cite{Quan(2021)}. Suppose that the line integral Lyapunov function $V_{\text{li}}\left(\mathbf{y}\right)$ is defined as (\ref{Vli0}). Then it is obtained that (i) $V_{\text{li}}\left(\mathbf{y} \right)>0$ if $\left \Vert \mathbf{y}\right \Vert \neq0$; (ii) if $ \left\Vert \mathbf{y}\right \Vert \rightarrow \infty,$ then $V_{\text{li}}\left(\mathbf{y}\right)\rightarrow \infty;$ (iii) if $V_{\text{li}}\left(\mathbf{y}\right)$ is bounded, then $\left \Vert \mathbf{y}\right \Vert $ is bounded.

\subsubsection{Two Smooth Functions}

Two smooth functions are defined for the design of Lyapunov-like barrier functions in our previous work \cite{Quan(2021)}. The first is
\begin{equation}
	\sigma \left(x,d_{1},d_{2}\right)\!=\!\left \{ 
	\begin{array}{c}
		1 \\ 
		Ax^{3}\!+\!Bx^{2}\!+\!Cx\!+\!D \\ 
		0%
	\end{array}%
	\right.\!
	\begin{array}{c}
		x\leq d_{1} \\ 
		d_{1}\leq x\leq d_{2} \\ 
		d_{2}\leq x%
	\end{array}
	\label{zerofunction}
\end{equation}
with $A=-2\left/\left(d_{1}-d_{2}\right)^{3}\right.,$ $B=3\left(d_{1}+d_{2}%
\right)\left/\left(d_{1}-d_{2}\right)^{3}\right.,$ $C=-6d_{1}d_{2}\left/
\left(d_{1}-d_{2}\right)^{3}\right.$, $D=d_{2}^{2}\left(3d_{1}-d_{2}%
\right)\left/\left(d_{1}-d_{2}\right)^{3}\right.$.
And the other is
\begin{equation}
	s\left(x,\epsilon_{\text{s}}\right)\!=\!\left \{ 
	\begin{array}{c}
		x \\ 
		\left(1\!-\!\epsilon_{\text{s}}\right)+\!\sqrt{\epsilon_{\text{s}%
			}^{2}\!-\!\left(\!x-x_{2}\!\right)^{2}} \\ 
		1%
	\end{array}%
	\right.%
	\begin{array}{c}
		0\leq x\leq x_{1} \\ 
		x_{1}\leq x\leq x_{2} \\ 
		x_{2}\leq x%
	\end{array}
	\label{sat}
\end{equation}
with $x_{2}=1+\frac{1}{\tan67.5^{\circ}}\epsilon_{\text{s}}$ and $%
x_{1}=x_{2}-\sin45^{\circ}\epsilon_{\text{s}}.$
\subsubsection{Single Panel Method}

\begin{figure}[h]
	\begin{centering}
		\includegraphics[scale=1]{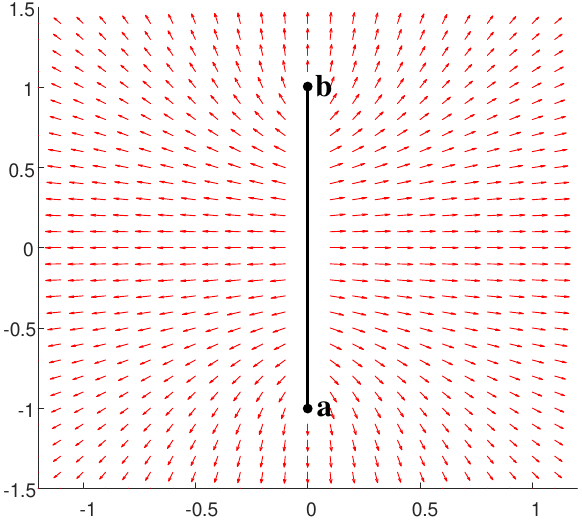}
		\par 
	\end{centering}
	\caption{Vector field of a single panel \cite{Kim(1992)}.}
	\label{PanelVF}
\end{figure}
Panel methods are widely used in aerodynamics calculations to obtain the solution to the potential flow problem around arbitrarily shaped bodies \cite{Kim(1992)}. Here the single panel method is used to represent the repulsive potential field of the tube boundary. Assume that there is a line segment $\left[\mathbf{a},\mathbf{b}\right]$, the potential at any point $\mathbf{p}$ induced by the sources contained within a small element $\text{d}x$ is shown as
\begin{align*}
	\text{d}\phi=\ln \left(\left \Vert \mathbf{p}-\left(\mathbf{a}+x\frac{\mathbf{b}-\mathbf{a}}{\left \Vert\mathbf{b}-\mathbf{a}\right \Vert}\right)\right \Vert-d\right)\text{d}x,
\end{align*}
where $d\geq0$ is the threshold distance. The induced repulsive potential function by the whole panel $\left[\mathbf{a},\mathbf{b}\right]$ is expressed as 
\begin{align}
	&\phi\left(\mathbf{p},\mathbf{a},\mathbf{b},d\right) \notag \\
	=&\int_{0}^{\left \Vert\mathbf{b}-\mathbf{a}\right \Vert}\ln \left(\left \Vert \mathbf{p}-\left(\mathbf{a}+x\frac{\mathbf{b}-\mathbf{a}}{\left \Vert\mathbf{b}-\mathbf{a}\right \Vert}\right)\right \Vert-d\right)\text{d}x. \label{PanelPF}
\end{align}
Given $\mathbf{a}=\left[0\ -1\right]^{\text{T}}$, $\mathbf{b}=\left[0\ 1\right]^{\text{T}}$, $d=0$, the corresponding negative vector field $-\partial \phi/\partial \mathbf{p}$ is shown in Fig. \ref{PanelVF}. It can be seen that the orientation is orthogonal to the line segment $\left[\mathbf{a},\mathbf{b}\right]$ when the point $\mathbf{p}$ locates at the line $y=0$. The orientation is parallel to $\left[\mathbf{a},\mathbf{b}\right]$ when the point $\mathbf{p}$ locates at the line $x=0$. As the potential function $\phi$ is smooth and differentiable, the orientation of the vector field also changes smoothly. This phenomenon is important for the proof of no deadlock in the following. 

\subsubsection{Error Definition}

Here two errors are defined. The first is the projection error between the $i$th agent and the finishing line $\mathcal{C}\left( {{\mathbf{p}}}_{\text{{fr}}}\right) $, namely
\begin{equation*}
	\tilde{\mathbf{{p}}}{_{\text{l,}i}}\triangleq \mathbf{P}_{\text{t}}\left(\mathbf{p}_{i}-{{\mathbf{p}}}_{\text{{fr}}}\right),
\end{equation*}
where $i=1,\cdots,M$ and the matrix $\mathbf{P}_{\text{t}}=\mathbf{P}_{\text{t}}^{\text{T}}=\mathbf{t}_{\text{c}}\mathbf{t}_{\text{c}}^{\text{T}}$ is a positive semi-definite projection matrix mapping a vector in the direction of $\mathbf{t}_{\text{c}}.$ The second is a position error between the $i$th and $j$th agent, which is shown as
\begin{equation*}
	\tilde{\mathbf{{p}}}{_{\text{m,}ij}}\triangleq \mathbf{p}_{i}-{{\mathbf{p}}%
		_{j}},
\end{equation*}
where $i,j=1,\cdots,M, i \neq j$. Then, according to (\ref{SingleIntegral}), the derivatives of these errors are shown as
\begin{align}
	\dot{\tilde{\mathbf{p}}}{_{\text{l,}i}}& =\mathbf{P}_{\text{t}}\mathbf{
		v}_{\text{c},i},  \label{lmodel} \\
	\dot{\tilde{\mathbf{{p}}}}{_{\text{m,}ij}}& =\mathbf{v}_{\text{c},i}-%
	\mathbf{v}_{\text{c},j}  \label{mmodel}.
\end{align}

\subsection{Lyapunov-Like Function Design and Analysis}

\subsubsection{Line Integral Lyapunov Function for Approaching Finishing Line}
Define a smooth curve $S_{\tilde{\mathbf{{p}}}{_{\text{l,}i}}}$ from $
\mathbf{0}$ to $\tilde{\mathbf{{p}}}{_{\text{l,}i}}$. The line
integral of $\text{sat}\left( \mathbf{x},{v_{\text{m},i}}\right) $ along $
S_{\tilde{\mathbf{{p}}}{_{\text{l,}i}}}$ is shown as%
\begin{equation}
	V_{\text{l},i}=\int_{S_{\tilde{\mathbf{{p}}}{_{\text{l,}i}}}}\text{sa}{%
		\text{t}}\left( k_{1}\mathbf{x},{v_{\text{m},i}}\right) ^{\text{T}}\text{d}%
	\mathbf{x}, \label{Vli}
\end{equation}
where $k_1>0$. From the definition, it is obtained that $V_{\text{l},i}\geq 0.$
According to the line integrals of vectors, the function (\ref{Vli}) is rewritten as \cite{Quan(2021)}
\begin{align}
	V_{\text{l},i}&=\int_{0}^{t}\text{sa}{\text{t}}\left( k_{1}\tilde{\mathbf{{p}}}{_{\text{l,}i}}\left( \tau \right) ,{v_{\text{m},i}}\right) ^{\text{T}}
	\dot{\tilde{\mathbf{{p}}}}{_{\text{l,}i}}\left( \tau \right) \text{d}%
	\tau  \notag \\
	&=\int_{0}^{t}\text{sa}{\text{t}}\left( k_{1}\tilde{\mathbf{{p}}}{_{\text{l,}i}}\left( \tau \right) ,{v_{\text{m},i}}\right) ^{\text{T}}
	\mathbf{P}_{\text{t}}\mathbf{v}_{\text{c},i}\left( \tau \right) \text{d}
	\tau. \label{Vli1}
\end{align}
The objective of the designed velocity command is to make $V_{\text{l},i}$ zero, which implies that $\left \Vert \tilde{\mathbf{{p}}}{_{\text{l,}i}}
\right \Vert=0$ according to (\ref{Vli1}), namely the 
$i$th agent arrives at the finishing line $\mathcal{C}\left( {{\mathbf{p}}_{\text{f}}}\right) $.

\subsubsection{Barrier Function for Avoiding Conflict among Agents}
According to two smooth functions introduced in \eqref{zerofunction} and \eqref{sat}, the barrier function for the $i$th agent to avoid conflict with the $j$th agent is defined as
\begin{equation}
	V_{\text{m},ij}=\frac{k_{2}\sigma _{\text{m}}\left( \left \Vert \tilde{\mathbf{{p}}}{_{\text{m,}ij}}\right \Vert \right) }{\left( 1+\epsilon _{\text{%
				m}}\right) \left \Vert \tilde{\mathbf{{p}}}{_{\text{m,}ij}}\right \Vert
		-2r_{\text{s}}s\left( \frac{\left \Vert \tilde{\mathbf{{p}}}{_{\text{m,}ij}}\right \Vert }{2r_{\text{s}}},\epsilon _{\text{s}}\right) },
	\label{Vmij}
\end{equation}
where $k_2,\epsilon _{\text{m}},\epsilon _{\text{s}}>0$. Based on the definitions of the safety area and the avoidance area, the smooth function $\sigma \left( \cdot \right) $ in (\ref{zerofunction}) is defined as $\sigma _{\text{m}}\left( x\right) \triangleq \sigma \left( x,2r_{\text{s}},r_{\text{a}}+r_{\text{s}}\right)$. The detailed properties of $V_{\text{m},ij}$ is presented
in our previous work \cite{Quan(2021)}. 
The objective of the designed velocity command is to make $V_{\text{m},ij}$ zero or as small as possible, which implies that $
\left \Vert \tilde{\mathbf{{p}}}{_{\text{m,}ij}}\right \Vert >2r_{\text{s}}{,}$ namely the $i$th agent will not conflict with the $j$th agent. Compared with other traditional potential field functions and barrier functions, our barrier function \eqref{Vmij} has a boarder domain of definition and more threshold distances.
\subsubsection{Barrier Function for Keeping within Virtual Tube}

\begin{figure}[h]
	\begin{centering}
		\includegraphics[scale=1]{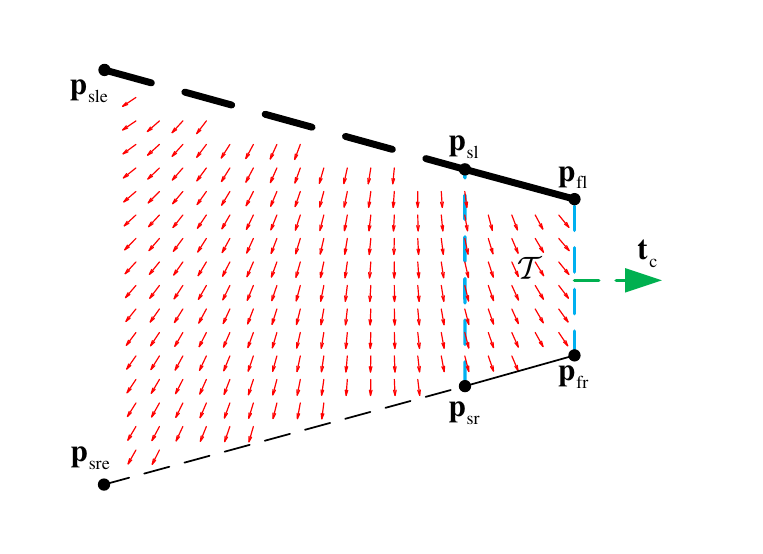}
		\par \end{centering}
	\caption{Vector field of the left extend tube boundary $\left[\mathbf{p}_\text{fl},\mathbf{p}_\text{sle}\right]$.}
	\label{LRVF}
\end{figure}

As shown in Fig. \ref{LRVF}, we first define two extended tube boundaries, $\left[\mathbf{p}_\text{fl},\mathbf{p}_\text{sle}\right]$ and $\left[\mathbf{p}_\text{fr},\mathbf{p}_\text{sre}\right]$, which satisfy 
\begin{align}
	&\left[\mathbf{p}_\text{fl},\mathbf{p}_\text{sl}\right] \subseteq \left[\mathbf{p}_\text{fl},\mathbf{p}_\text{sle}\right],  \label{DirL1} \\
	&\left[\mathbf{p}_\text{fr},\mathbf{p}_\text{sr}\right] \subseteq \left[\mathbf{p}_\text{fr},\mathbf{p}_\text{sre}\right], \label{DirR1}
\end{align}  
namely the line segments $\left[\mathbf{p}_\text{fl},\mathbf{p}_\text{sle}\right]$, $\left[\mathbf{p}_\text{fr},\mathbf{p}_\text{sre}\right]$ are longer than $\left[\mathbf{p}_\text{fl},\mathbf{p}_\text{sl}\right]$, $\left[\mathbf{p}_\text{fr},\mathbf{p}_\text{sr}\right]$, respectively. According to the potential function of the single panel (\ref{PanelPF}), two barrier functions for the $i$th agent to keep within the virtual tube are defined as
\begin{align}
	V_{\text{tl},i}&=k_{3}\phi\left(\mathbf{p}_i,\mathbf{p}_\text{sle},\mathbf{p}_\text{fl},r_\text{s}\right), \label{Vtli} \\ 	V_{\text{tr},i}&=k_{3}\phi\left(\mathbf{p}_i,\mathbf{p}_\text{sre},\mathbf{p}_\text{fr},r_\text{s}\right), \label{Vtri}
\end{align}
where $k_3>0$. For deadlock avoidance, the chosen of points $\mathbf{p}_\text{sle}$, $\mathbf{p}_\text{sre}$ must meet the following requirements
\begin{align}
	-\mathbf{t}_{\text{c}}^{\text{T}}\left(\frac{\partial V_{\text{tl},i}}{\partial \mathbf{p}_i }\right)^{\text{T}}&\geq0, \mathbf{p}_i \in \mathcal{T},  \label{DirL2}\\
	-\mathbf{t}_{\text{c}}^{\text{T}}\left(\frac{\partial V_{\text{tr},i}}{\partial \mathbf{p}_i }\right)^{\text{T}}&\geq0, \mathbf{p}_i \in \mathcal{T}  \label{DirR2}.
\end{align}
As shown in Fig. \ref{LRVF}, the constraints \eqref{DirL2}, \eqref{DirR2} imply that the angles between negative gradient directions of $V_{\text{tl},i}$, $V_{\text{tr},i}$ inside the virtual tube  $\mathcal{T}$ and the moving direction $\mathbf{t}_{\text{c}}$ must keep smaller than $90^{\circ}$, which plays a crucial role in the stability proof. It is obvious that if line segments $\left[\mathbf{p}_\text{fl},\mathbf{p}_\text{sle}\right]$, $\left[\mathbf{p}_\text{fr},\mathbf{p}_\text{sre}\right]$ are long enough, the constraints \eqref{DirL2}, \eqref{DirR2} are always satisfied.

The objective of the designed velocity command is to make $V_{\text{tl},i}$ and $V_{\text{tr},i}$ as small as possible, which implies that 
$\text{dist}\left(\mathbf{p}_i,\left[\mathbf{p}_\text{fl},\mathbf{p}_\text{sle}\right]\right)>r_\text{s}$ and $\text{dist}\left(\mathbf{p}_i,\left[\mathbf{p}_\text{fr},\mathbf{p}_\text{sre}\right]\right)>r_\text{s}$, where the function $\text{dist}\left(\cdot\right)$ is defined as the Euclidean distance, namely the $i$th agent will keep within the virtual tube. 

\subsection{Distributed Swarm Controller}

\begin{figure}[h]
	\begin{centering}
		\includegraphics[scale=1.1]{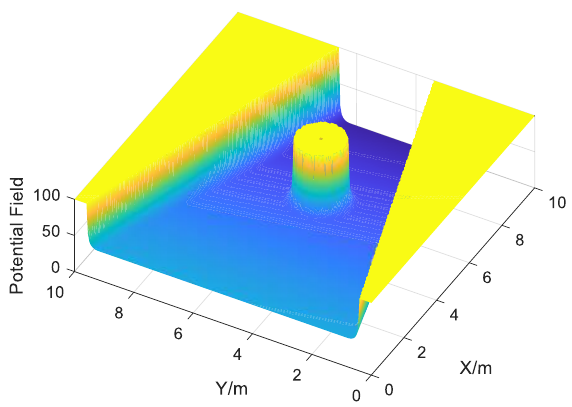}
		\par \end{centering}
	\caption{Potential field of the trapezoid virtual tube. There are some small modifications on the value, whose purpose is to make the figure more intuitive.}
	\label{APFplot}
\end{figure}

The velocity command of the $i$th agent is designed as
\begin{align}
	\mathbf{v}_{\text{c},i}
	=&-\text{sat}\left( \underset{\text{Line
			Approaching}}{\underbrace{\mathbf{P}_{\text{t}}\text{sat}\left( {{k}_{1}}\tilde{\mathbf{{p}}}{_{\text{l,}i}},{v_{\text{m},i}}\right) }}+{\underset{\text{Agent
				Avoidance}}{\underbrace{\underset{j\in \mathcal{N}_{\text{m},i}}{\overset{}{%
						\sum }}-b_{ij}\tilde{\mathbf{{p}}}{_{\text{m,}ij}}}}}\right. \notag\\
	&\left.+\underset{\text{Virtual Tube Keeping}}{\underbrace{\left(\frac{\partial V_{\text{tl},i}}{\partial \mathbf{p}_i }\right)^{\text{T}}+\left(\frac{\partial V_{\text{tr},i}}{\partial \mathbf{p}_i }\right)^{\text{T}}}},{v_{\text{m},i}}\right),  \label{controller1}
\end{align}
where $b_{ij} =-\frac{\partial V_{\text{m},ij}}{\partial \left \Vert \tilde{\mathbf{{p}}}{_{\text{m,}ij}}\right \Vert }\frac{1}{\left \Vert \tilde{\mathbf{{p}}}{_{\text{m,}ij}}\right \Vert }. $ The controller \eqref{controller1} is a distributed swarm control form and can work autonomously without wireless communication. The ID of any neighboring agent is not required.  Consider a scenario that an agent is moving within a trapezoid virtual tube, in the middle of which there exists the other agent. Fig. \ref{APFplot} shows the potential field of this trapezoid virtual tube. It can be observed that the value of the potential field near the line $x=0\text{m}$ is larger than the value near the line $x=10\text{m}$. Besides, the positions near the tube boundary and the other agent have very large values of the potential field. Hence, the agent with its initial position at the line $x=0\text{m}$ will ``slide down to''
the line $x=10\text{m}$ meanwhile avoiding collision with the other agent and keeping moving inside the virtual tube.

\subsection{Stability Analysis}

In order to investigate the stability of the proposed controller, a function is defined as follows
\begin{equation*}
	{V}=\underset{i=1}{\overset{M}{\sum }}\left( V_{\text{l},i}+\frac{1}{2}%
	\underset{j=1,j\neq i}{\overset{M}{\sum }}V_{\text{m},ij}+V_{\text{tl},i}+V_{\text{tr},i}\right) ,
\end{equation*}
where $V_{\text{l},i},V_{\text{m},ij},V_{\text{tl},i}, V_{\text{tr},i}$ are defined in (\ref{Vli1}), (\ref{Vmij}), (\ref{Vtli}), (\ref{Vtri}) respectively. According to (\ref{lmodel}), (\ref{mmodel}), the derivative of ${V}$ is shown as
\begin{align*}
	{\dot{V}}&=\sum_{i=1}^{M}\left( \text{sat}\left( {{k}%
		_{1}}\tilde{\mathbf{p}}{_{\text{l,}i}},{v_{\text{m},i}}\right) ^{\text{%
			T}}\mathbf{P}_{\text{t}}\mathbf{v}_{\text{c},i}\right.\\
	&\left.-\frac{{1}}{2}\sum_{j=1,j\neq i}^{M}b_{ij}\tilde{\mathbf{p}}_{\text{m,}ij}^{\text{T}}\left( \mathbf{v}_{\text{c},i}-\mathbf{v}_{\text{c},j}\right) +\frac{\partial V_{\text{tl},i}}{\partial \mathbf{p}_i } \mathbf{v}_{\text{c},i}+\frac{\partial V_{\text{tr},i}}{\partial \mathbf{p}_i } \mathbf{v}_{\text{c},i}\right)
	\\
	&=\sum_{i=1}^{M}\left( \mathbf{P}_{\text{t}}\text{sat}\left( {{k}_{1}}\tilde{\mathbf{p}}{_{\text{l,}i}},{v_{\text{m},i}}\right)+\sum_{j\in\mathcal{N}_{\text{m},i}}-b_{ij}\tilde{\mathbf{p}}_{\text{m,}ij}\right.\\
	&\left.+\left(\frac{\partial V_{\text{tl},i}}{\partial \mathbf{p}_i }\right)^{\text{T}}+\left(\frac{\partial V_{\text{tl},i}}{\partial \mathbf{p}_i }\right)^{\text{T}}\right) ^{\text{T}}\mathbf{v}_{\text{c},i}.
\end{align*}
With the definitions of $\mathcal{N}_{\text{m},i}$ and $V_{\text{m},ij}$, there exists $\sum_{j=1,j\neq i}^{M}b_{ij}\tilde{\mathbf{p}}_{\text{m,}ij}=\sum_{j\in\mathcal{N}_{\text{m},i}}b_{ij}\tilde{\mathbf{p}}_{\text{m,}ij}$.
By using the velocity command (\ref{controller1}) for all agents, ${\dot{V}}$ satisfies ${\dot{V}}\leq 0$.

Before introducing the main result, an important lemma is needed.

\textbf{Lemma 2}. \cite{Quan(2021)} Under \textit{Assumptions 1, 2}, suppose that the velocity command is designed as {(\ref{controller1}). }Then
there exist sufficiently small $\epsilon _{\text{m}},\epsilon _{\text{s}}>0$
in $b_{ij}$ such that $%
\left \Vert \tilde{\mathbf{{p}}}{_{\text{m,}ij}}\left( t\right)
\right
\Vert >2r_{\text{s}},$ $\text{dist}\left(\mathbf{p}_i\left( t\right),\left[\mathbf{p}_\text{fl},\mathbf{p}_\text{sl}\right]\right)>r_\text{s}$, $\text{dist}\left(\mathbf{p}_i\left( t\right),\left[\mathbf{p}_\text{fr},\mathbf{p}_\text{sr}\right]\right)>r_\text{s}$, $t\in \lbrack 0,\infty )$ for all ${{%
		\mathbf{p}}_{i}(0)}$, $i,j=1,\cdots ,M,i\neq j.$


With \emph{Lemmas 1, 2} in hand, the main result is stated as follows. 

\textbf{Theorem 1}. Under \textit{Assumptions 1, 2}, suppose that
(i) the velocity command is designed as  (\ref{controller1});  (ii) given ${\epsilon }_{\text{0}}>0{,}$ if (\ref{arrivialairway}) is satisfied, then $b_{ij}\equiv 0$, $\left(\partial V_{\text{tl},i}/\partial \mathbf{p}_i\right)^{\text{T}}\equiv \mathbf{0}$, $\left(\partial V_{\text{tr},i}/\partial \mathbf{p}_i\right)^{\text{T}}\equiv \mathbf{0}$, which implies that the $i$th agent is removed from the virtual tube mathematically. Then, given ${\epsilon }_{\text{0}}>0$, there
exist sufficiently small $\epsilon _{\text{m}},\epsilon _{\text{s}}>0$
in $b_{ij}$ and $t_{1}>0$ such that all
agents can satisfy (\ref{arrivialairway}) at $t\geq t_{1},$ meanwhile ensuring  $%
\left \Vert \tilde{\mathbf{{p}}}{_{\text{m,}ij}}\left( t\right)
\right
\Vert >2r_{\text{s}},$ $\text{dist}\left(\mathbf{p}_i\left( t\right),\left[\mathbf{p}_\text{fl},\mathbf{p}_\text{sl}\right]\right)>r_\text{s}$, $\text{dist}\left(\mathbf{p}_i\left( t\right),\left[\mathbf{p}_\text{fr},\mathbf{p}_\text{sr}\right]\right)>r_\text{s}$, $t\in \lbrack 0,\infty )$ for all ${{%
		\mathbf{p}}_{i}(0)}$, $i,j=1,\cdots ,M,i\neq j$.

\emph{Proof}. See \emph{Appendix}. $\square$

\subsection{Modified Distributed Swarm Controller}

The controller (\ref{controller1}) has two apparent imperfections in use. 
\begin{itemize}[leftmargin=*]
	\item The first problem is that any agent can approach the finishing line but its speed will slow down to zero. The reason is that $\tilde{\mathbf{{p}}}{_{\text{l,}i}}=\mathbf{0}$ when $\mathbf{p}_{i}$ locates on $\mathcal{C}\left( {{\mathbf{p}}_{\text{f}}}\right)$. 
	\item The second problem is that the values of $\mathbf{p}_\text{sle}$, $\mathbf{p}_\text{sre}$ are difficult to obtain. The specific mathematical forms of $\partial V_{\text{tl},i}/\partial \mathbf{p}_i$ and $\partial V_{\text{tr},i}/\partial \mathbf{p}_i$ are also very complicated and inconvenient for practical use.
\end{itemize}
 
To solve the first problem, we define a modified finishing line $\mathcal{C}\left( {{\mathbf{p}}}_{\text{{fr}}}^{\prime }\right) $ as shown in Fig. {\ref{airwaytube}}, denoted by  
$
	\mathcal{C}\left( {{\mathbf{p}}}_{\text{{fr}}}^{\prime }\right) =\left \{ \mathbf{x}\in {{\mathbb{R}}^{2}}:\mathbf{t}_{\text{c}}^{\text{T}}\left(\mathbf{x}-{{\mathbf{p}}}_{\text{{fr}}}^{\prime }\right) =0\right \},
$
where ${{\mathbf{p}}}_{\text{{fr}}}^{\prime }={{\mathbf{p}}}_{\text{{fr}}}+\gamma \mathbf{t}_{\text{c}}, \gamma=\frac{\max\left(v_{\text{m},i}\right)}{k_1}$. In this case,  the line approaching term becomes
\begin{align*}
	&\quad\ \mathbf{P}_{\text{t}}\text{sa}{\text{t}}\left( {{k}_{1}}\mathbf{P}_{\text{t}%
	}\tilde{\mathbf{{p}}}{_{\text{l,}i}},{v_{\text{m},i}}\right) \\
	&=\mathbf{P}_{\text{t}}\text{sa}{\text{t}}\left({{k
		}_{1}}\mathbf{P}_{\text{t}}\left( \mathbf{p}_{i}-{{\mathbf{p}}}_{%
		\text{{fr}}}\right)-\max \left( {v_{%
			\text{m},i}}\right)  \mathbf{t}_{\text{c}},{v_{\text{m},i}}\right)  =-v_{\text{m},i}\mathbf{t}_{\text{c}}.
\end{align*}

\begin{figure}[h]
	\begin{centering}
		\includegraphics[scale=1]{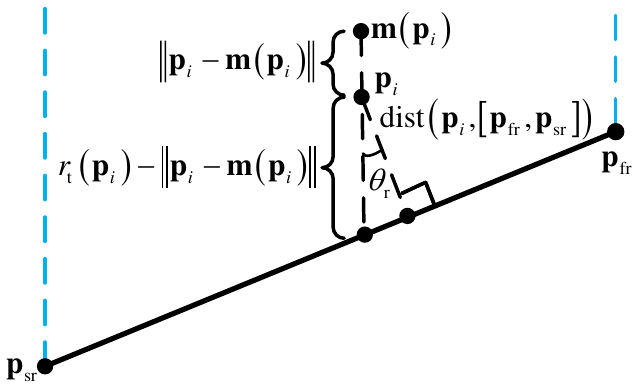}
		\par \end{centering}
	\caption{The reason for proposing the revised safety radius.}
	\label{Threeaera2}
\end{figure}

To solve the second problem, a non-potential term is introduced to approximate the performance of  $\partial V_{\text{tl},i}/\partial \mathbf{p}_i$ and $\partial V_{\text{tr},i}/\partial \mathbf{p}_i$. An Euclidean distance error is defined between the $i$th agent and the tube boundary, which is shown as 
\begin{equation*}
	d{_{\text{t,}i}} \triangleq r_{\text{t}}\left( \mathbf{p}_{i}\right) - \left \Vert \mathbf{p}_{i}-\mathbf{m}\left( \mathbf{p}_i\right) \right \Vert.
\end{equation*}
The derivative of this error is shown as 
\begin{equation*}
	\dot{d }{_{\text{t,}i}}\!=\!\left(\frac{\partial r_{\text{t}}\left( \mathbf{p}_{i}\right)}{\partial \mathbf{p}_{i}}-\frac{\left(\mathbf{p}_{i}-\mathbf{m}\left( \mathbf{p}_i\right)\right)^{\text{T}}}{\left \Vert \mathbf{p}_{i}-\mathbf{m}\left( \mathbf{p}_i\right) \right \Vert}\left( \mathbf{I}_{\text{2}}-\frac{\partial \mathbf{m}\left( \mathbf{p}_i\right)}{\partial  \mathbf{p}_i}\right) \right)
	\mathbf{v}_{\text{c},i}.
\end{equation*}
For ensuring $\mathcal{S}_{i}\cap \partial \mathcal{T}=\varnothing$, at least 
$d{_{\text{t,}i}} >r_{\text{s}}$ is required. However, this
constraint is not enough because the real distance from $\mathbf{p}_i$ to $\partial \mathcal{T}$, which may be $\text{dist}\left(\mathbf{p}_i,\left[\mathbf{p}_\text{fr},\mathbf{p}_\text{sr}\right]\right)$, is usually smaller than $d{_{\text{t,}i}}$ as shown in Fig. \ref{Threeaera2}. Hence, we put forward a revised safety radius and have the following proposition.

\textbf{Proposition 1}. For any
$\mathbf{p}_i\in \mathcal{T},$ if and only if $	d{_{\text{t,}i}}>r_{\text{s}}^{\prime },$ then $%
\mathcal{S}_{i}\cap \partial \mathcal{T}=\varnothing$. The constant $r_{\text{s}
}^{\prime }$ is the \emph{revised safety radius}, which is defined as
\begin{equation}
	r_{\text{s}}^{\prime }=\frac{r_{\text{s}}}{\min \left( -\mathbf{t}_{\text{c}%
		}^{\text{T}}\frac{{{\mathbf{p}}_{\text{sl}}-{\mathbf{p}}_{\text{fl}}}}{\left
			\Vert {{\mathbf{p}}_{\text{sl}}-{\mathbf{p}}_{\text{fl}}}\right \Vert },-%
		\mathbf{t}_{\text{c}}^{\text{T}}\frac{{{\mathbf{p}}_{\text{sr}}}-\mathbf{p}_{%
				\text{fr}}}{\left \Vert {{\mathbf{p}}_{\text{sr}}-{\mathbf{p}}_{\text{fr}}}%
			\right \Vert }\right) }.  \label{rs'}
\end{equation}

\textit{Proof}. Define $\theta _{\text{r}}$ as the angle between the line $\overline{{{\mathbf{p}}_{\text{sr}}}\mathbf{p}_{\text{fr}}}$ and the vector $-%
\mathbf{t}_{\text{c}}$. And $\theta _{\text{l}}$ is the angle between the line $\overline{{{\mathbf{p}}_{\text{sl}}{\mathbf{p}}_{\text{fl}}%
}}$ and the vector $-\mathbf{t}_{\text{c}}$. These two angles satisfy $	\cos \theta _{\text{r}} =-\mathbf{t}_{\text{c}}^{\text{T}}\frac{{{\mathbf{p}}_{\text{sr}}}-\mathbf{p}_{\text{fr}}}{\left \Vert {{\mathbf{p}}_{\text{sr}}-{\mathbf{p}}_{\text{fr}}}\right \Vert }$ and $\cos \theta _{\text{l}} =-\mathbf{t}_{\text{c}}^{\text{T}}\frac{{{\mathbf{p}}_{\text{sl}}-{\mathbf{p}}_{\text{fl}}}}{\left \Vert {{\mathbf{p}}_{\text{sl}}-{\mathbf{p}}_{\text{fl}}}\right \Vert }$.
For any $\mathbf{p}_{i}\in \mathcal{T},$ as shown in Fig. \ref%
{Threeaera2}, the distance from $\mathbf{p}_{i}$\ to $\partial 
\mathcal{T}$ is shown as
$\text{dist}\left( \mathbf{p}_{i},\partial \mathcal{T}\right) = \cos \theta _{\text{l}} \left(	r_{\text{t}}\left( \mathbf{p}_{i}\right) - \left \Vert \mathbf{p}_{i}-\mathbf{m}\left( \mathbf{p}_i\right) \right \Vert \right)$
or $\text{dist}\left( \mathbf{p}_{i},\partial \mathcal{T}\right) =\cos \theta _{\text{r}}\left(	r_{\text{t}}\left( \mathbf{p}_{i}\right) - \left \Vert \mathbf{p}_{i}-\mathbf{m}\left( \mathbf{p}_i\right) \right \Vert \right) $.
If and only if $\mathcal{S}_{i}\cap \partial \mathcal{T}=\varnothing,$ then
dist$\left( \mathbf{p}_i,\partial \mathcal{T}\right) >r_{\text{s}}.$ $\square $

Then the barrier function for the $i$th agent to keep within the virtual tube is defined as
\begin{equation}
	V_{\text{t},i}=\frac{k_{3}\sigma _{\text{t}}\left(d{_{\text{t,}i}}\right) }{\left( 1+\epsilon _{\text{t}}\right) d{_{\text{t,}i}}
		-r_{\text{s}}^\prime s\left( \frac{d{_{\text{t,}i}} }{r_{\text{s}}^\prime},\epsilon _{\text{s}}\right) }, \label{Vti}
\end{equation}
where $\epsilon _{\text{t}}>0$. Here the smooth function $\sigma \left( \cdot \right) $ in (\ref{zerofunction}) is defined as $\sigma _{\text{t}}\left( x\right) \triangleq \sigma \left( x,r_{\text{s}}^{\prime },r_{\text{a}}\right) $ .
The function $V_{\text{t},i}$ has similar properties to $V_{\text{m},ij}$.
The objective of the designed velocity command is to make $V_{\text{t},i}$ zero or as small as possible. This implies that $d{_{\text{t,}i}} >r_{\text{s}}^{\prime }$, namely the $i$th agent will keep within the virtual tube. 

\begin{figure}[h]
	\begin{centering}
		\includegraphics[scale=1]{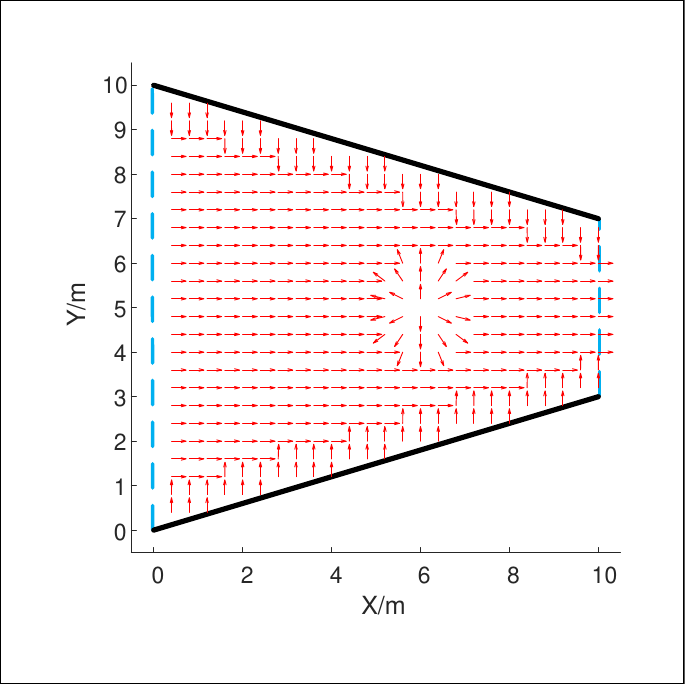}
		\par \end{centering}
	\caption{Vector field of a trapezoid virtual tube with the modified controller \eqref{controller2}.}
	\label{VFplot}
\end{figure}

Let $\mathbf{p}$ be the collection $\left( \mathbf{p}_{1},\cdots ,\mathbf{p}_{M}\right) $. Then the modified distributed swarm controller is shown as 
\begin{align}
	\mathbf{v}_{\text{c},i}&\!=\!\mathbf{v}\left( \mathcal{T},\mathbf{p}_{i},
	\mathbf{p},r_\text{s}^{\prime }\right) \label{controller2}\\
	&\!=\!-\text{sat}\left( -v_{\text{m},i}\mathbf{t}_{\text{c}}-\sum_{j\in \mathcal{N}_{\text{m},i}}b_{ij}\tilde{\mathbf{{p}}}{_{\text{m,}ij}}+\left(\mathbf{I}_2-\mathbf{P}_\text{t}\right)\mathbf{c}_i,{v_{\text{m},i}}\right) \notag,
\end{align}
where $\left(\mathbf{I}_2-\mathbf{P}_\text{t}\right)\mathbf{c}_i$ is the modified virtual tube keeping term and $\mathbf{c}_i$ is expressed as
\begin{equation*}
	\mathbf{c}_{i}\!=\!\frac{\partial V_{\text{t},i}}{\partial d{_{\text{t,}i}}}\left(\frac{\partial r_{\text{t}}\left( \mathbf{p}_{i}\right)}{\partial \mathbf{p}_{i}}-\frac{\left(\mathbf{p}_{i}-\mathbf{m}\left( \mathbf{p}_i\right)\right)^{\text{T}}}{\left \Vert \mathbf{p}_{i}-\mathbf{m}\left( \mathbf{p}_i\right) \right \Vert}\left( \mathbf{I}_{\text{2}}-\frac{\partial \mathbf{m}\left( \mathbf{p}_i\right)}{\partial  \mathbf{p}_i}\right) \right)^{\text{T}}. 
\end{equation*}
Consider a scenario that an agent is moving within a trapezoid virtual tube, in the middle of which there exists another agent. Fig. \ref{VFplot} shows the vector field of this trapezoid virtual tube with the modified swarm controller \eqref{controller2}.

\textbf{Remark 2}. The term $\mathbf{c}_{i}$ is the gradient of $V_{\text{t},i}$, namely $\mathbf{c}_{i}=\left(\partial V_{\text{t},i}/\partial \mathbf{p}_i\right)^{\text{T}}$, which is orthogonal to the line 
$\overline{{{\mathbf{p}}_{\text{sr}}}\mathbf{p}_{\text{fr}}}$ or $\overline{{{\mathbf{p}}_{\text{sl}}}\mathbf{p}_{\text{fl}}}$. And $\left(\mathbf{I}_2-\mathbf{P}_\text{t}\right)\mathbf{c}_i$ is a non-potential velocity command component, which is always orthogonal to $\mathbf{t}_{\text{c}}$. To avoid deadlock, directly applying $\mathbf{c}_{i}$ in (\ref{controller1}) is not feasible. Hence the use of the single panel method is necessary. Suppose that there exists 
$\left[\mathbf{p}_\text{fl},\mathbf{p}_\text{sl}\right] \subset \left[\mathbf{p}_\text{fle},\mathbf{p}_\text{sle}\right]$ and $\left[\mathbf{p}_\text{fr},\mathbf{p}_\text{sr}\right] \subset \left[\mathbf{p}_\text{fre},\mathbf{p}_\text{sre}\right]$, the orientation changes of $\partial V_{\text{tl},i}/\partial \mathbf{p}_i$ and $\partial V_{\text{tr},i}/\partial \mathbf{p}_i$ inside the virtual tube may be negligibly small if $\left[\mathbf{p}_\text{fle},\mathbf{p}_\text{sle}\right]$ and $\left[\mathbf{p}_\text{fre},\mathbf{p}_\text{sre}\right]$ are long enough. Hence, we can choose appropriate points $\mathbf{p}_\text{fle}$, $\mathbf{p}_\text{sle}$, $\mathbf{p}_\text{fre}$, $\mathbf{p}_\text{sre}$ so that 
the orientations of $\partial V_{\text{tl},i}/\partial \mathbf{p}_i$ and $\partial V_{\text{tr},i}/\partial \mathbf{p}_i$ are both approximately orthogonal to $\mathbf{t}_{\text{c}}$, which explains why $\left(\mathbf{I}_2-\mathbf{P}_\text{t}\right)\mathbf{c}_i$ can approximate $\partial V_{\text{tl},i}/\partial \mathbf{p}_i$ and $\partial V_{\text{tr},i}/\partial \mathbf{p}_i$. Hence, \emph{Theorem 1} also remains valid if its condition (i) is replaced with the velocity command designed as  (\ref{controller2}).

\textbf{Remark 3}. Compared with $V_{\text{tl},i}$, $V_{\text{tr},i}$ in (\ref{Vtli}), (\ref{Vtri}), the barrier function $V_{\text{t},i}$ in (\ref{Vti}) has its unique advantage of the broader application. In practice, the case such as $\text{dist}\left(\mathbf{p}_i,\partial \mathcal{T}\right)<r_\text{s}$
may still happen in practice due to unpredictable uncertainties. Under this circumstance, the potential functions $V_{\text{tl},i}$ and $V_{\text{tr},i}$
have computation errors, while $V_{\text{t},i}$ still works well and the modified keeping term $\left(\mathbf{I}_2-\mathbf{P}_\text{t}\right)\mathbf{c}_i$ dominates the velocity command $\mathbf{v}\left( \mathcal{T},\mathbf{p}_{i},
\mathbf{p},r_\text{s}^{\prime }\right) $, which implies that $	d{_{\text{t,}i}}$ will be increased very fast so that the $i$th agent can keep away from the tube boundary immediately.

\section{Distributed Control for Passing a Connected Quadrangle Virtual Tube}


\subsection{Control Aeras of a Quadrangle}

\begin{figure}[!t]
	\centerline{\includegraphics[width=\columnwidth]{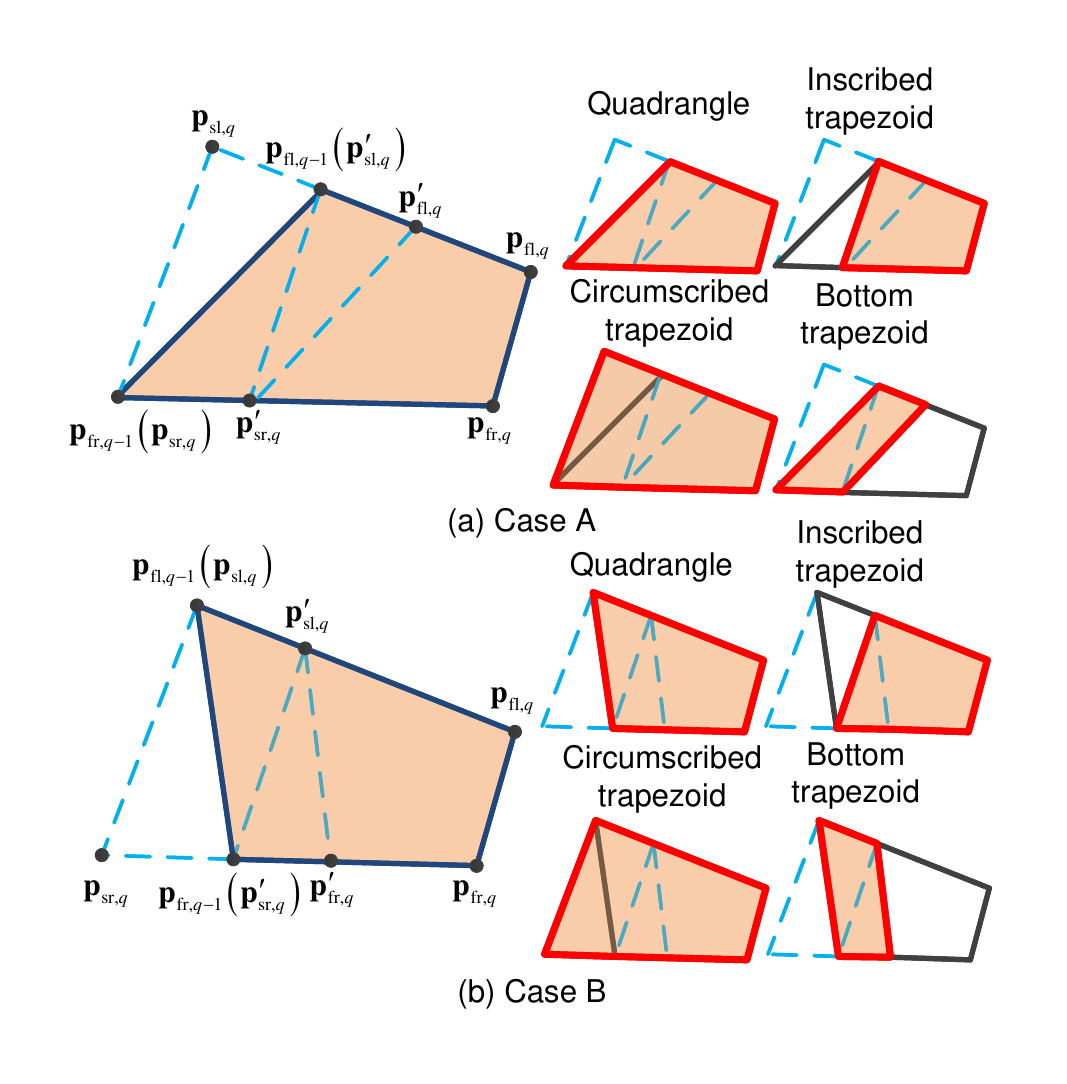}}
	\caption{Inscribed trapezoid, circumscribed trapezoid and bottom trapezoid
	of a quadrangle.}
\label{inscribedcircumscribed}
\end{figure}

As in (\ref{ConnectedQuadrangle}), a connected quadrangle virtual tube ${\mathcal{Q}}$ is composed of $N$ single quadrangles. In (\ref{controller2}), the distributed swarm controller for a trapezoid virtual tube has been proposed. However, this controller is not suitable for an arbitrary quadrangle, of which any pairs of edges may not be parallel. Hence it is necessary to transform a quadrangle to several trapezoids.

To design the swarm controller for a connected quadrangle virtual tube, three areas are defined in a general quadrangle. First, for the $q$th quadrangle, the \emph{inscribed trapezoid} $\mathcal{T}_{\mathcal{Q}_{q}\text{i}}$ and the \emph{circumscribed trapezoid} $\mathcal{T}_{\mathcal{Q}_{q}\text{c}}$ are defined as
$\mathcal{T}_{\mathcal{Q}_{q}\text{i}}=\mathcal{T}_{\mathcal{Q}_{q}^{\prime}}\cap \mathcal{T}_{\mathcal{Q}_{q}^{\prime \prime }}$, $	\mathcal{T}_{\mathcal{Q}_{q}\text{c}}=\mathcal{T}_{\mathcal{Q}_{q}^{\prime}}\cup \mathcal{T}_{\mathcal{Q}_{q}^{\prime \prime }},$
where
\begin{align*}
	\mathcal{T}_{\mathcal{Q}_{q}^{\prime }} =&\left \{  \mathbf{x}\in {{%
			\mathbb{R}}^{2}}: \mathbf{n}_{\text{l,}q}^{\text{T}}\left( 
	\mathbf{x}-{{\mathbf{p}}_{\text{fl,}q}}\right) \geq 0,\mathbf{n}_{\text{r,}q}^{\text{T}}\left( \mathbf{x}-{{\mathbf{p}}_{\text{fr,}q}}\right) \geq 0, \right. \\
	&\left.-\mathbf{t}_{\text{c,}q}^{\text{T}}\left( \mathbf{x}-{{\mathbf{p}}_{\text{fr,}q}}\right) \geq 0,\mathbf{t}_{\text{c,}q}^{\text{T}}\left( \mathbf{x}-{{
			\mathbf{p}}_{\text{fl,}q-1}}\right) \geq 0\right \}, 
\end{align*}
\begin{align}
	\mathcal{T}_{\mathcal{Q}_{q}^{\prime \prime }} =&\left \{ \mathbf{x}
	\in {{\mathbb{R}}^{2}}: \mathbf{n}_{\text{l,}q}^{\text{T}}\left( 
	\mathbf{x}-{{\mathbf{p}}_{\text{fl,}q}}\right) \geq 0,\mathbf{n}_{\text{r,}%
		q}^{\text{T}}\left( \mathbf{x}-{{\mathbf{p}}_{\text{fr,}q}}\right) \geq 0,\right. \notag\\
	&\left.-\mathbf{t}_{\text{c,}q}^{\text{T}}\left( \mathbf{x}-{{\mathbf{p}}_{\text{fr,}%
			q}}\right) \geq 0,\mathbf{t}_{\text{c,}q}^{\text{T}}\left( \mathbf{x}-{{%
			\mathbf{p}}_{\text{fr,}q-1}}\right) \geq 0\right \} .  \label{TQ}
\end{align}
It is easy to see that $\mathcal{T}_{\mathcal{Q}_{q}\text{i}}\subseteq \mathcal{Q}_{q}\subseteq 
\mathcal{T}_{\mathcal{Q}_{q}\text{c}}.$ For example, as shown in Fig. {\ref{inscribedcircumscribed}, }the
inscribed trapezoid $\mathcal{T}_{\mathcal{Q}_{q}\text{i}}$ and the circumscribed trapezoid $\mathcal{T}_{\mathcal{Q}_{q}\text{c}}$ of the quadrangle $
\mathcal{Q}_{q}$ are
$\mathcal{T}_{\mathcal{Q}_{q}\text{i}} =\mathcal{T}\left( {{\mathbf{p}}_{\text{fr,}q},{\mathbf{p}}_{\text{fl,}q},{\mathbf{p}}}_{\text{sl,}q}^{\prime
},{{\mathbf{p}}}_{\text{sr,}q}^{\prime }\right) $, $\mathcal{T}_{\mathcal{Q}_{q}\text{c}} =\mathcal{T}\left( {{\mathbf{p}}_{\text{fr,}q},{\mathbf{p}}_{\text{fl,}q},{\mathbf{p}}}_{\text{sl,}q},{{\mathbf{p}}}_{\text{sr,}q}\right)$.
Then, the \emph{bottom trapezoid} $\mathcal{T}_{\mathcal{Q}_{q}\text{b}}$ of $\mathcal{Q}_{q}$ is defined as
\begin{align*}
	\mathcal{T}_{\mathcal{Q}_{q}\text{b}} =&\left \{  \mathbf{x}\in {{
			\mathbb{R}}^{2}}: \mathbf{n}_{\text{l,}q}^{\text{T}}\left( 
	\mathbf{x}-{{\mathbf{p}}_{\text{fl,}q}}\right) \geq 0,\mathbf{n}_{\text{r,}%
		q}^{\text{T}}\left( \mathbf{x}-{{\mathbf{p}}_{\text{fr,}q}}\right) \geq
	0\right \} \\
	&\cap \left( \left( \mathcal{A}\cup \mathcal{B}\right) -\left( \mathcal{A}\cap \mathcal{B}\right) \right),
\end{align*}
where $\mathcal{A}=\left \{ \mathbf{x}\in {{\mathbb{R}}^{2}}:\mathbf{t}_{\text{c,}q-1}^{\text{T}}\left( \mathbf{x}-{{\mathbf{p}}}_{\text{sl,}q}^{\prime }\right) \geq 0\right \}$, $\mathcal{B}=\left \{  \mathbf{x}\in {{\mathbb{R}}^{2}}:\mathbf{t}_{\text{c,}q-1}^{\text{T}}\left( \mathbf{x}-{{\mathbf{p}}}_{\text{sr,}q}^{\prime }\right) \geq 0\right \}$.
As shown in Fig. {\ref{inscribedcircumscribed}, }$\mathcal{T}_{\mathcal{Q}%
	_{q}\text{b}}$ is a trapezoid with a base $\left[ {{\mathbf{p}}_{\text{%
			fr,}q-1},{\mathbf{p}}_{\text{fl,}q-1}}\right] $ and a diagonal $\left[ 
{{\mathbf{p}}}_{\text{sl,}q}^{\prime },{{\mathbf{p}}}_{\text{sr,}q}^{\prime }%
\right] ,$ where $\left[ {{\mathbf{p}}}_{\text{sl,}q}^{\prime },{{\mathbf{p}}%
}_{\text{sr,}q}^{\prime }\right] $ is further a base of $\mathcal{T}_{%
	\mathcal{Q}_{q}\text{i}}$. Obviously, there exists $\mathcal{T}_{\mathcal{Q}_{q}\text{b}%
}\subset \mathcal{Q}_{q}.$ Then a following proposition is proposed.

\textbf{Proposition 2}. If ${{\mathbf{t}}_{\text{c,}q}}={{\mathbf{t}}_{\text{%
			c,}q-1},}$ then there exists $\mathcal{Q}_{q}=\mathcal{T}_{\mathcal{Q}_{q}\text{i}}=%
\mathcal{T}_{\mathcal{Q}_{q}\text{c}}$ and $\mathcal{T}_{\mathcal{Q}_{q}%
	\text{b}}=\varnothing .$

\textit{Proof}. If ${{\mathbf{t}}_{\text{c,}q}}={{\mathbf{t}}_{\text{c,}q-1}}
$, then $\mathcal{Q}_{q}$ is expressed as
\begin{align*}
	\mathcal{Q}_{q}^{\prime }=&\left \{ \mathbf{x}\in {{\mathbb{R}}^{2}}
	: \mathbf{n}_{\text{l,}q}^{\text{T}}\left( \mathbf{x}-{{\mathbf{p}%
		}_{\text{fl,}q}}\right) \geq 0,\mathbf{n}_{\text{r,}q}^{\text{T}}\left( 
	\mathbf{x}-{{\mathbf{p}}_{\text{fr,}q}}\right) \geq 0,\right. \notag\\
	&\left.-\mathbf{t}_{\text{c,}
		q}^{\text{T}}\left( \mathbf{x}-{{\mathbf{p}}_{\text{fr,}q}}\right) \geq 0,%
	\mathbf{t}_{\text{c,}q}^{\text{T}}\left( \mathbf{x}-{{\mathbf{p}}_{\text{fl,}%
			q-1}}\right) \geq 0\right \}
\end{align*}%
or 
\begin{align*}
	\mathcal{Q}_{q}^{\prime\prime}=&\left \{  \mathbf{x}\in {{\mathbb{R}}%
		^{2}}: \mathbf{n}_{\text{l,}q}^{\text{T}}\left( \mathbf{x}-{{%
			\mathbf{p}}_{\text{fl,}q}}\right) \geq 0,\mathbf{n}_{\text{r,}q}^{\text{T}%
	}\left( \mathbf{x}-{{\mathbf{p}}_{\text{fr,}q}}\right) \geq 0,\right. \notag\\
	&\left.-\mathbf{t}_{\text{c,}q}^{\text{T}}\left( \mathbf{x}-{{\mathbf{p}}_{\text{fr,}q}}\right)
	\geq 0,\mathbf{t}_{\text{c,}q}^{\text{T}}\left( \mathbf{x}-{{\mathbf{p}}_{%
			\text{fr,}q-1}}\right) \geq 0\right \} .
\end{align*}%
Recalling {(\ref{TQ})}, there exists $\mathcal{Q}_{q}=\mathcal{T}_{\mathcal{Q}_{q}^{\prime }}$, $\mathcal{Q}_{q}=\mathcal{T}_{\mathcal{Q}_{q}^{\prime \prime }}.$
Consequently, it is obtained that
${{\mathbf{p}}_{\text{fl,}q-1}}={{\mathbf{p}}}_{\text{sl,}q}^{\prime }={{\mathbf{p}}}_{\text{sl,}q}$, ${{\mathbf{p}}_{\text{fr,}q-1}}={{\mathbf{p}}}_{\text{sr,}q}^{\prime }={{\mathbf{p}}}_{\text{sr,}q}$.
This implies that the points ${{\mathbf{p}}}_{\text{sl,}q}^{\prime }$ and ${{\mathbf{p}}}_{%
	\text{sr,}q}^{\prime }$ are on the line $\overline{ {{\mathbf{p}}_{\text{fl,}q-1},{%
			\mathbf{p}}_{\text{fr,}q-1}}}.$ Since $\mathbf{t}_{\text{c,}q}={{%
		\mathbf{t}}_{\text{c,}q-1},}$ we further have $\mathcal{A}=\mathcal{B}.$ Therefore, $\left( \mathcal{A}\cup \mathcal{B}\right) -\left( \mathcal{A}\cap \mathcal{B}\right) =\varnothing.$
Thus it is obtained that $\mathcal{T}_{\mathcal{Q}_{q}\text{b}}=\varnothing .$ $\square $

\subsection{A Direct Switching Logic for Moving across Two Quadrangles}

\begin{figure}[h]
	\begin{centering}
		\includegraphics[scale=1]{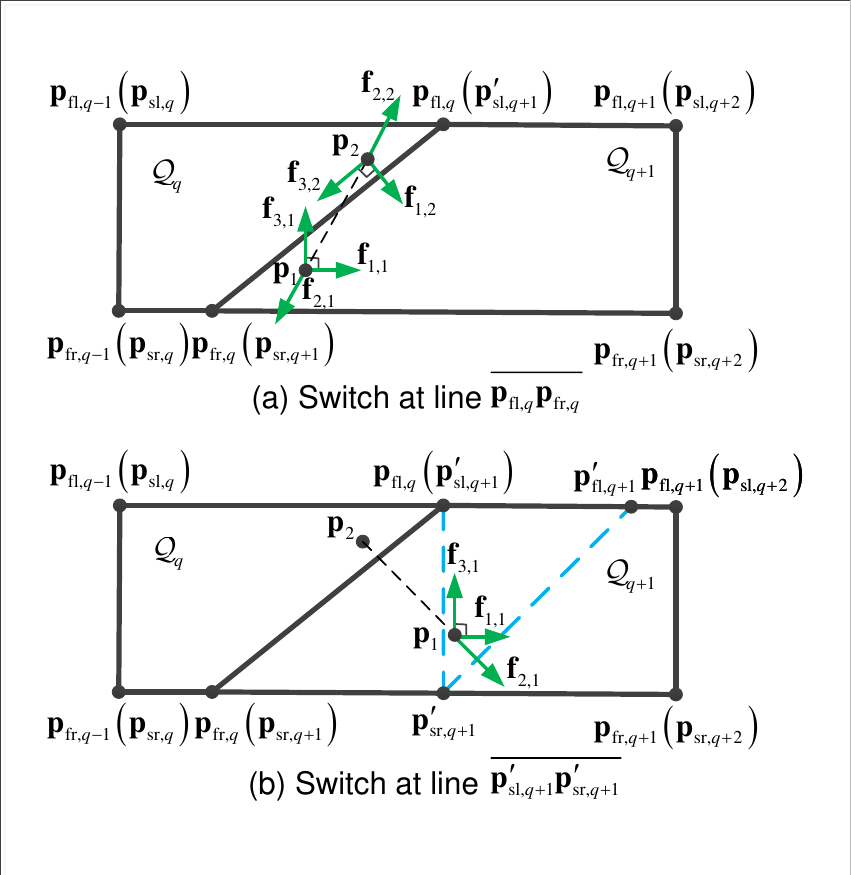}
		\par \end{centering}
	\caption{The difference between the direct switching logic and the modified switching logic.}
	\label{problem}
\end{figure}
When an agent moves across the finishing line of the quadrangle $\mathcal{Q}_{q}$, it will enter the next quadrangle $\mathcal{Q}_{q+1}$. So it is necessary to build up a control switching logic between two adjoint quadrangles. First a direct and straightforward
switching logic is designed as
\begin{equation}
	\mathbf{v}_{\text{c,}i}=\mathbf{v}\left( \mathcal{T}_{\mathcal{Q%
		}_{q}\text{c}},\mathbf{p}_{i},\mathbf{p},r_{\text{s},q}^{\prime }\right) \text{ if }\mathbf{p}_{i} \in \mathcal{Q}_{q}  \label{directlogic}
\end{equation}
where 
\begin{equation*}
	r_{\text{s,}q}^{\prime } =\frac{r_{\text{s}}}{\min \left(-\mathbf{t}_{\text{c,}q}^{\text{T}}\frac{{{\mathbf{p}}_{\text{sl,}q}-{\mathbf{p}}_{\text{fl,}q}}}{\left \Vert {{\mathbf{p}}_{\text{sl,}q}-{\mathbf{p}}_{\text{fl,}q}}\right \Vert },-\mathbf{t}_{\text{c,}q}^{\text{T}}\frac{{{\mathbf{p}}_{\text{sr,}q}}-\mathbf{p}_{\text{fr,}q}}{\left \Vert {{\mathbf{p}}_{\text{sr,}q}-{\mathbf{p}}_{\text{fr,}q}}\right \Vert }\right) }
\end{equation*}
and $i=1,\cdots ,M$, $q=1,\cdots ,N$.
This switching logic implies that if $\mathbf{p}_{i}$ is in the $q$th quadrangle, the $i$th agent will adopt the distributed controller (\ref{controller2}) to pass through its corresponding circumscribed trapezoid virtual tube $\mathcal{T}_{\mathcal{Q}_{q}\text{c}}.$ As there exists $\mathcal{Q}_{q}\subseteq 
\mathcal{T}_{\mathcal{Q}_{q}\text{c}}$, the controller \eqref{controller2} can be reused in $\mathcal{Q}_{q}$. However, there are two problems appearing at the connection between two quadrangles to be solved. 

The first problem is the \emph{deadlock}. For the sake of simplicity, here the velocity commands are described as forces. As shown in Fig. \ref{problem}, for the $i$th agent in the $q+1$th quadrangle, ${{\mathbf{f}}_{\text{1},i}}$ is the attractive force from the finishing line $
\left[ {{\mathbf{p}}_{\text{fr,}q+1},{\mathbf{p}}_{\text{fl,}q+1}}\right]$, ${{\mathbf{f}}_{\text{2},i}}$
is the repulsive force from other agents, and ${{\mathbf{f}}_{\text{3},i}}$ is the repulsive force from
the tube boundary, namely
${{\mathbf{f}}_{\text{1},i}}=-k_1{\kappa}_{v_{\text{m},i}}\mathbf{P}_\text{t}\mathbf{\tilde{p}}_\text{l,i}^{\prime}$, ${{\mathbf{f}}_{\text{2},i}}=\sum_{j=1,j \neq i}^{M}{b}_{ij}{\mathbf{\tilde{p}}}_{\text{m,}ij}$, ${{\mathbf{f}}_{\text{3},i}}=-\left(\mathbf{I}_{2}-\mathbf{P}_\text{t}\right)\mathbf{c}_{i}$, where  $i,j=1,\cdots,M, i \neq j$. It should be noted that ${{\mathbf{f}}_{\text{1},i}}$ is always perpendicular to ${{\mathbf{f}}_{\text{3},i}}$. As shown in Fig. \ref{problem}(a), if the $1$st agent is the first to arrive at $\left[{{\mathbf{p}}_{\text{fr,}q},{\mathbf{p}}_{\text{fl,}q}}\right]$ and the $2$nd agent is the second, then the $1$st agent will switch its controller according to (\ref{directlogic}). The resultant forces on both agents may be ${{\mathbf{f}}_{\text{1},1}+\mathbf{f}_{\text{2},1}+\mathbf{f}_{\text{3},1}}
=\mathbf{0}$, ${{\mathbf{f}}_{\text{1},2}+\mathbf{f}_{\text{2},2}+\mathbf{f}_{\text{3},2}}
=\mathbf{0}$. The reason for this phenomenon is that the $2$nd agent
is closer to $\left[ {{\mathbf{p}}_{\text{fr,}q+1},{\mathbf{p}}_{\text{fl,}q+1}}%
\right] $ than the $1$st agent at the moment, although the $1$st agent is
closer to $\left[ {{\mathbf{p}}_{\text{fr,}q},{\mathbf{p}}_{\text{fl,}q}}%
\right] $ than the $2$nd agent. Under this circumstance, the deadlock will happen. The second problem is that the agent may move outside the $q+1$th quadrangle virtual tube once the agent just enters this quadrangle virtual tube because $\left[ {{\mathbf{p}}_{\text{fr,}q-1},{\mathbf{p%
	}}_{\text{fr,}q}}\right] $, $\left[ {{\mathbf{p}}_{\text{fr,}q},{\mathbf{p}%
	}_{\text{fr,}q+1}}\right]$ or $\left[ {{\mathbf{p}}_{\text{fl,}q-1},{\mathbf{p%
	}}_{\text{fl,}q}}\right] $, $\left[ {{\mathbf{p}}_{\text{fl,}q},{\mathbf{p}%
	}_{\text{fl,}q+1}}\right]$ have different slopes.

\subsection{A Modified Switching Logic for Moving across Two Quadrangles}

To solve the two problems proposed in the last subsection, a modified switching logic is designed as
\begin{equation}
	\mathbf{v}_{\text{c,}i}\!=\! \left \{ 
	\begin{aligned}
	\mathbf{v}\left(\! \mathcal{T}_{\mathcal{Q}_{q}\text{b}},\!\mathbf{p}_{i},\mathbf{p}, \max \left(\! r_{\text{s,}q-1}^{\prime },r_{\text{s,}%
	q}^{\prime }\!\right) \!\right) &\  \mathbf{p}_{i}\!\in\! \mathcal{Q}_{q}\!-\!
		\mathcal{T}_{\mathcal{Q}_{q}\text{i}} \\
		\mathbf{v}\left(\! \mathcal{T}_{\mathcal{Q}_{q}\text{i}},\!\mathbf{p}_{i},\mathbf{p}, \max \left(\! r_{\text{s,}q}^{\prime },r_{\text{s,}%
			q+1}^{\prime }\!\right) \!\right) & \  \mathbf{p}_{i}\!\in\! \mathcal{T}_{%
			\mathcal{Q}_{q}\text{i}} \\
	\end{aligned}
	\right .  \label{switching}
\end{equation}
where $i=1,\cdots ,M$, $q=1,\cdots ,N$ and $r_{\text{s},0}^{\prime } =r_{\text{s},N+1}^{\prime } =0.$ As shown in Fig. \ref{problem}(b), when the $1$st agent locates in $\mathcal{Q}_{q}-
\mathcal{T}_{\mathcal{Q}_{q}\text{i}}$, there is no possibility of deadlock as the attractive force and repulsive force from the tube boundary of the $1$st and $2$nd agents have the same directions. Then the following theorem is proposed.

\textbf{Theorem 2}. Under \textit{Assumptions 1, 2}, suppose that (i) the velocity command is designed as (\ref{switching}); (ii) given ${\epsilon }_{\text{0}}>0{,}$ if (\ref{arrivialairway}) is satisfied, $b_{ij}\equiv 0$ and $\mathbf{c}_{i}\equiv \mathbf{0}$, which implies that the $i$th agent is removed from the virtual tube mathematically; (iii) there exists $r_{\text{t}}\left( \mathcal{T}_{\mathcal{Q}_{q}\text{c}}\right) >\max \left( r_{\text{s,}q-1}^{\prime },r_{\text{s,}q}^{\prime },r_{\text{s,}q+1}^{\prime }\right) ,q=1,\cdots ,N$. Then, given ${\epsilon}_{\text{0}}>0$, there exist sufficiently small $\epsilon _{\text{m}},\epsilon_{\text{s}}>0$ in $b_{ij}$, $\epsilon _{\text{t}}>0$ in $\mathbf{c}_{i}$ and $t_{1}>0$ such that all agents  satisfy (\ref{arrivialairway}) at $t\geq t_{1},$ meanwhile ensuring $\mathcal{S}_{i}\left(t\right)\cap  \mathcal{S}_{j}\left(t\right)=\varnothing $, $\mathcal{S}_{i}\left(t\right)\cap \partial\mathcal{Q}=\varnothing$, $t\in \lbrack 0,\infty )$ for all ${{\mathbf{p}}_{i}(0)}$, $i,j=1,\!\cdots\! ,M,i\neq\! \!j$.

\textit{Proof}. Similarly to the proof of \textit{Theorem 1}, any agent, which arrives at $\left[ {{\mathbf{p}}}_{\text{fl,}q},{{\mathbf{p}}}_{\text{fr,}q}\right] $, can pass through $\mathcal{T}_{\mathcal{Q}_{q+1}\text{b}}$ by
$
	\mathbf{v}_{\text{c,}i}=\mathbf{v}\left( \mathcal{T}_{\mathcal{Q%
		}_{q+1}\text{b}},\mathbf{p}_{i},\mathbf{p},\max \left( r_{\text{s,}%
		q}^{\prime },r_{\text{s,}q+1}^{\prime }\right) \right) .
$
Since $\max \left( r_{\text{s,}q}^{\prime },r_{\text{s,}q+1}^{\prime
}\right) $ is adopted, this agent will keep within $\mathcal{T}_{\mathcal{%
		Q}_{q+1}\text{b}}.$ As $ \left[{{\mathbf{p}}}_{\text{sl,}q+1}^{\prime },{{%
		\mathbf{p}}}_{\text{sr,}q+1}^{\prime }\right]\subset \mathcal{T}_{\mathcal{Q}%
	_{q+1}\text{b}},$ there exists a time that one of the agents, saying the $1$st agent,
will arrive at $\left[{\mathbf{p}}_{\text{sl,}q+1}^{\prime },{{\mathbf{p}}}
_{\text{sr,}q+1}^{\prime }\right].$ According to {(\ref{switching})}, its controller will
switch to 
$
	\mathbf{v}_{\text{c,}1}=\mathbf{v}\left( \mathcal{T}_{\mathcal{Q%
		}_{q+1}\text{i}},\mathbf{p}_{1},\mathbf{p}, \max \left( r_{\text{s,}%
		q+1}^{\prime },r_{\text{s,}q+2}^{\prime }\right) \right) .
$
As the agent model \eqref{SingleIntegral} is a single-integrator, there is no transition process, namely the switching logic \eqref{switching} has no influence on the Lyapunov analysis.
Since $\max \left( r_{\text{s,}q+1}^{\prime },r_{\text{s,}q+2}^{\prime
}\right) $ is adopted, the $1$st agent will keep within $\mathcal{T}_{\mathcal{%
		Q}_{q+1}\text{i}}.$ Condition (iii) is a necessary condition to show that the
circumscribed trapezoid of a quadrangle is wide enough for at least one
agent to pass.  Hence, the $1$st agent can arrive at $\left[ {{\mathbf{p}}}_{\text{fl,%
	}q+1},{{\mathbf{p}}}_{\text{fr,}q+1}\right]$ by \textit{Theorem 1} and all agents can pass through$
\mathcal{T}_{\mathcal{Q}_{q+1}}$. If the agent arriving at $\left[ {{%
		\mathbf{p}}}_{\text{fl,}q+1},{{\mathbf{p}}}_{\text{fr,}q+1}\right] $ has no
effect on the agents behind, we can repeat the analysis to conclude this proof,
where condition (ii) is used to analyze $\mathcal{Q}_{N}$ as there is no next quadrangle virtual tube anymore. $\square $

\section{Simulations and Experiments}
Simulations and experiments are given to show the effectiveness of the proposed method. A video about simulations and experiments is available on
\href{https://youtu.be/S04n-BMikfM}{https://youtu.be/S04n-BMikfM}.

\subsection{Numerical Simulation with Different Maximum Velocities}
\begin{figure}[!t]
	\centerline{\includegraphics[width=\columnwidth]{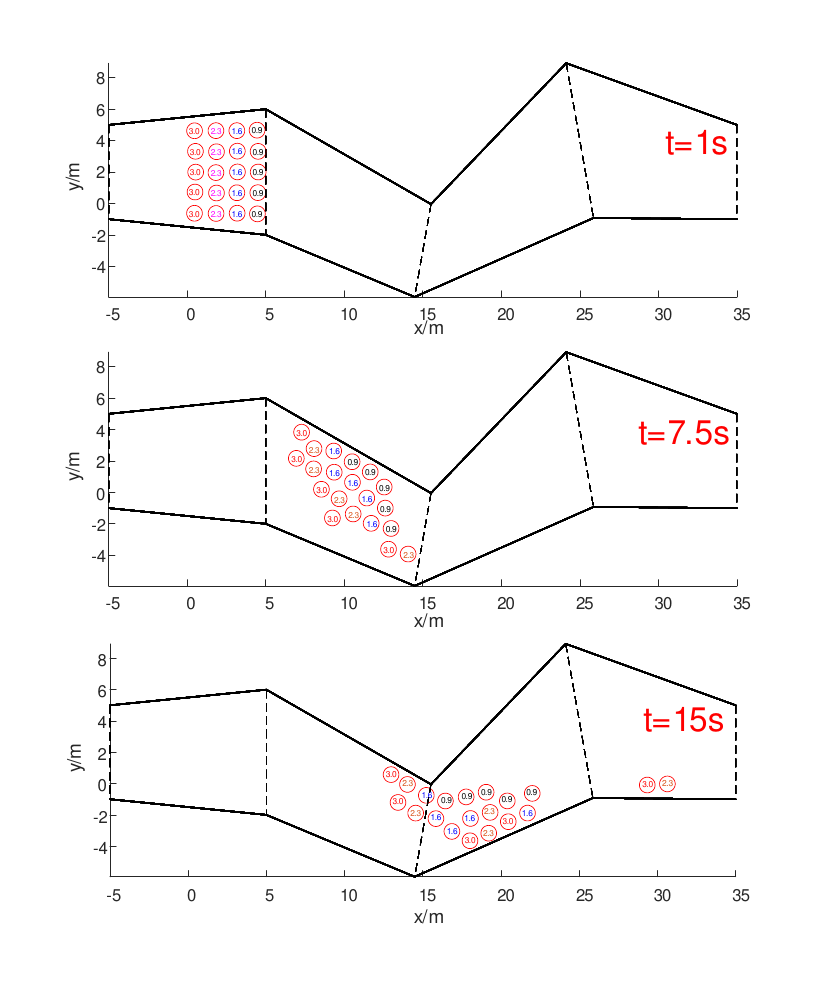}}
	\caption{Simulation snapshot.}
	\label{simulation1}
\end{figure}

\begin{figure}[!t]
	\centerline{\includegraphics[width=\columnwidth]{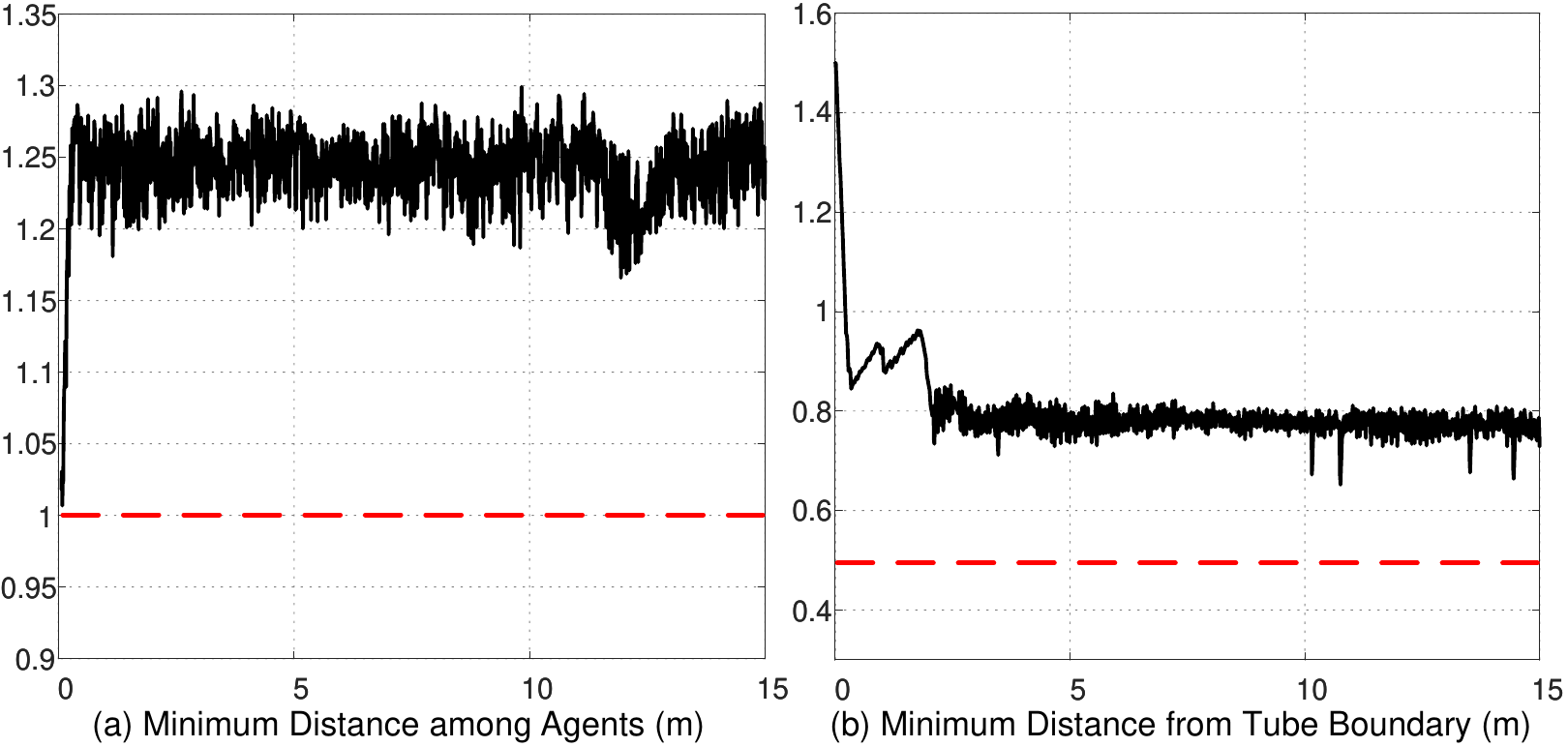}}
	\caption{Minimum distance among agents and minimum distance from the tube boundary in the simulation.}
\label{simulation2}
\end{figure}


The validity and feasibility of the proposed method is numerically verified in a simulation. The simulation is implemented on Matlab 2021a, Windows 10, Intel(R) Core(TM) i7-8700, 32GB DDR4 2666MHz. The simulation step is 0.001s. Consider a scenario that one multi-agent system composed of $M=20$ agents passes through a predefined connected quadrangle virtual tube. All the agents satisfy the control model in (\ref{SingleIntegral}). The swarm controller in (\ref{switching}) is applied to guide this multi-agent system. The connected quadrangle virtual tube is set as shown in Fig. \ref{simulation1}, where the first quadrangle $\mathcal{Q}_{1}$ is a trapezoid.
The parameters and initial conditions of the simulation are set as follows. The control parameters are $k_2=k_3=1$, $\epsilon_{\text{m}}=\epsilon_{\text{t}}=\epsilon_{\text{s}}=10^{-6}$.
All agents with the safety radius $r_\text{s} = 0.5\text{m}$, the avoidance radius $r_\text{a}= 0.8\text{m}$ and initial speeds being zero are arranged symmetrically in a rectangular space in the beginning. As shown in Fig. \ref{simulation1}, the boundaries of the safety area are represented by red circles. To show the ability to control different types of agents at the same time with our proposed method, we set the agents' maximum speed to four different constants.
The corresponding maximum speed for each agent is shown with different colors in the center of the safety area. 

The simulation lasts 15 seconds and three snapshots are shown in Fig. \ref{simulation1}. It can be seen that the agents with the largest speed $v_{\text{m},i}=3.0\text{m/s}$ are in the last column in the beginning. Then they have the trend to overtake other agents ahead. In the whole process, agents can change their relative positions freely instead of maintaining a fixed geometry structure. It is clear from Fig. \ref{simulation2}(a) that the minimum distance between any two agents is always larger than $2r_\text{s}=1\text{m}$, which implies that there is no collision in the swarm. In Fig. \ref{simulation2}(b), the minimum distance from the tube boundary among all agents keeps larger than $r_\text{s}=0.5\text{m}$ all the time. Therefore, the agents can avoid colliding with each other and keep moving in the connected quadrangle virtual tube under the swarm controller (\ref{switching}). Besides, the average calculation time of our controller \eqref{switching} is 0.001737s. The same simulation is implemented with the CBF method \cite{Wang(2017)} for comparison. The average calculation time of the CBF method is 0.05725s. It can be observed that our method has a high computational efficiency.

\subsection{Experiment}
\begin{figure}[!t]
	\centerline{\includegraphics[width=\columnwidth]{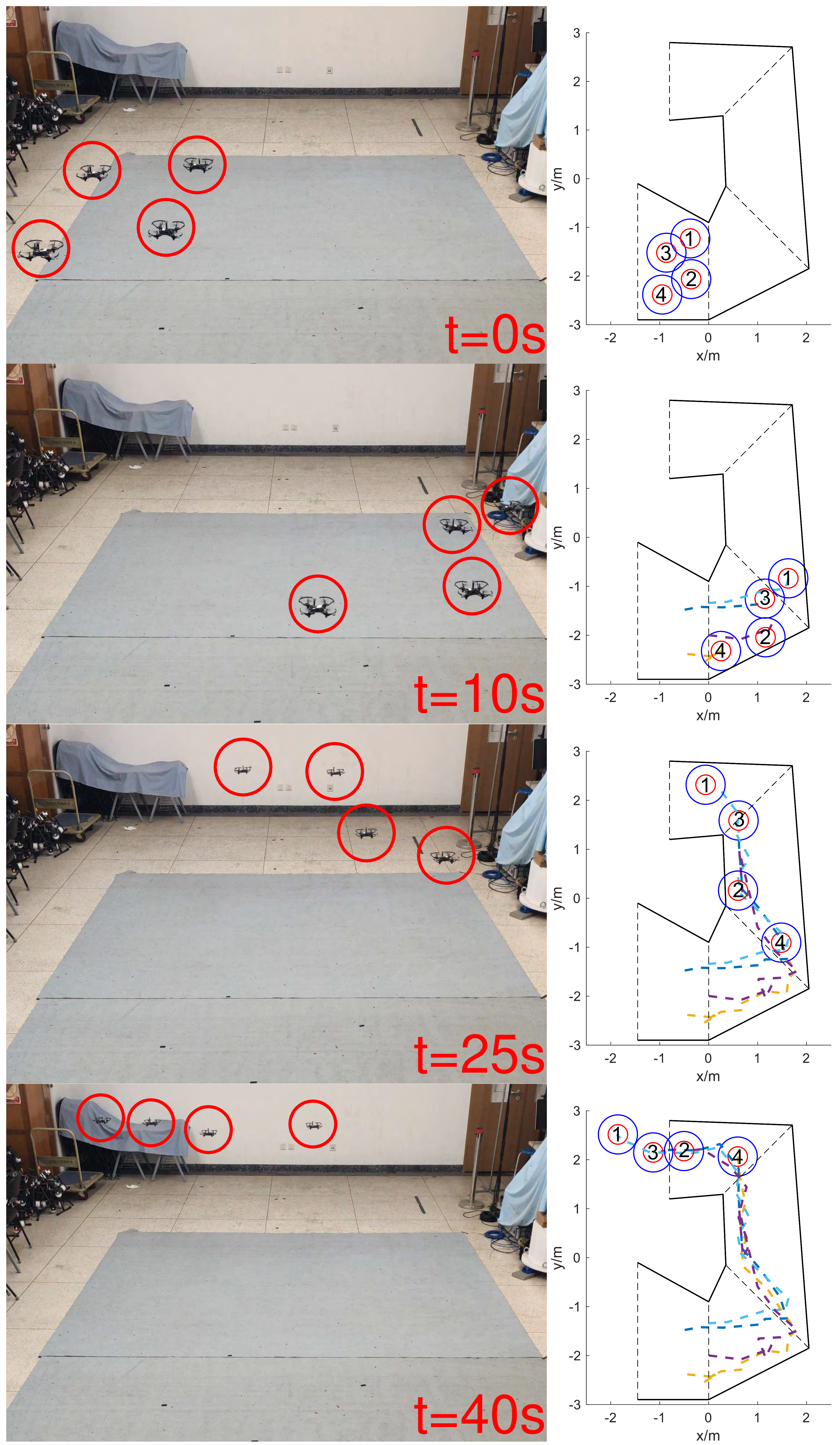}}
	\caption{Experiment snapshot.}
	\label{experiment1}
\end{figure}

\begin{figure}[!t]
	\centerline{\includegraphics[width=\columnwidth]{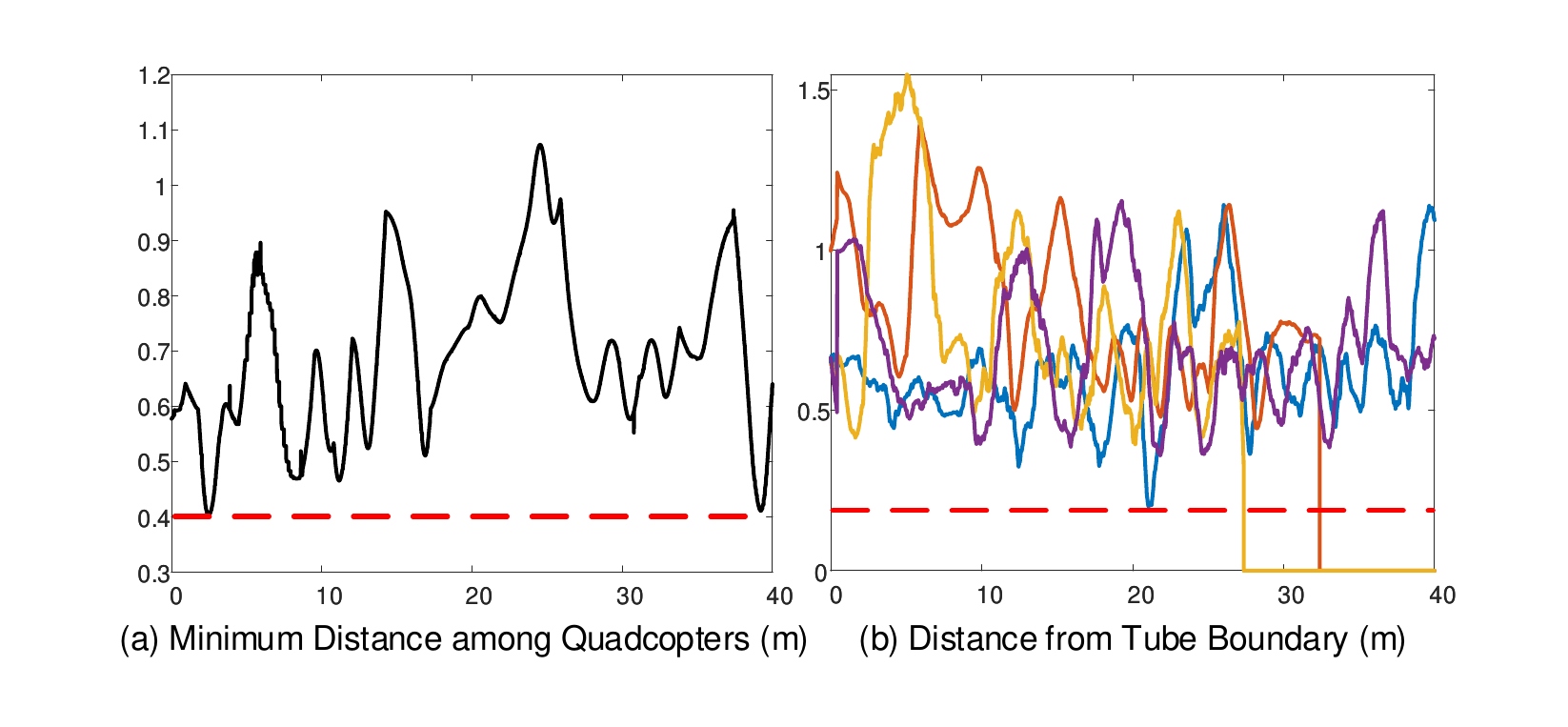}}
	\caption{Minimum distance among quadcopters and distances from the tube boundary in the experiment. Two curves in plot (b) approaching zero means that their corresponding quadcopters, namely the 1st and the 3rd quadcopters, have passed the finishing line of the last quadrangle virtual tube, which can be confirmed in the snapshot of $t=40\text{s}$ in Fig. \ref{experiment1}. }
	\label{experiment3}
\end{figure}
A real experiment is carried out in a laboratory room with $M=4$ Tello quadcopters and an OptiTrack motion capture system, which provides the positions and orientations of all quadcopters. A laptop computer is connected to Tello quadcopters and OptiTrack with a local wireless network, running the proposed distributed controller (\ref{switching}). The connected quadrangle virtual tube consists of four parts, in which the first one is a trapezoid virtual tube. The control parameters are $k_2=k_3=1$, $\epsilon_{\text{m}}=\epsilon_{\text{t}}=\epsilon_{\text{s}}=10^{-6}$.
All quadcopters have the safety radius $r_\text{s} = 0.2\text{m}$, the avoidance radius $r_\text{a} = 0.4\text{m}$ and the maximum speed $v_{\text{m},i}=0.4\text{m/s},i=1,\cdots,4$. As shown in Fig. \ref{experiment1}, the boundaries of the safety and avoidance area are represented by red and blue circles respectively. It can be observed that the initial positions of all quadcopters are at the first trapezoid virtual tube with initial speeds being zero.

The experiment lasts 40 seconds and four snapshots are shown in Fig. \ref{experiment1}. As same as the numerical simulation, these quadcopters can change their relative positions freely instead of maintaining a fixed geometry structure. It is clear from Fig. \ref{experiment3}(a) that the minimum distance between any two quadcopters  is always larger than $2r_\text{s}=0.4\text{m}$, which implies that there is no collision among quadcopters. In Fig. \ref{experiment3}(b), the distances from the tube boundary of all quadcopters keep larger than $r_\text{s}=0.2\text{m}$ when the quadcopters are inside the virtual tube.



\section{Conclusion}
The single trapezoid virtual tube passing problem and the connected quadrangle virtual tube passing problem are proposed and then solved in this paper. Based on the artificial potential field method with a control saturation, the distributed swarm controller is finally proposed for multiple agents to pass through a connected quadrangle virtual tube. Lyapunov-like functions are designed elaborately, and formal analysis and proofs are made to show that the virtual tube passing problem can
be solved, namely all agents avoid collision with each other and keep within the virtual tube in \emph{Lemma 1}, and all agents pass through the virtual tube without getting trapped in \emph{Theorems 1, 2}. Simulations and experiments are given to show the effectiveness and performance of the proposed method in different kinds of conditions. The focus of future work will be on the connected quadrangle virtual tube planning problem. The controller proposed in this paper has solved the passing problem under any circumstance. However, the passing efficiency is not mentioned, which is closely related to the virtual tube planning. Obviously, an appropriate planning can bring great improvement in passing efficiency. Besides, the condition that there exist obstacles inside the virtual tube should also be considered.

\appendices

\section{Proof of Theorem 1}
According to \textit{Lemma 2}, the agents are able to avoid conflict with each other and keep within the trapezoid virtual tube, namely $\left \Vert \tilde{\mathbf{{p}}}{_{\text{m,}ij}}\left( t\right)\right\Vert >2r_{\text{s}},$ $\text{dist}\left(\mathbf{p}_i\left( t\right),\left[\mathbf{p}_\text{fl},\mathbf{p}_\text{sl}\right]\right)>r_\text{s}$,  $\text{dist}\left(\mathbf{p}_i\left( t\right),\left[\mathbf{p}_\text{fr},\mathbf{p}_\text{sr}\right]\right)>r_\text{s}$, $t\in \lbrack 0,\infty )$ for all ${{\mathbf{p}}_{i}(0)}$, $i,j=1,\cdots ,M,i\neq j$. In the following, the reason why the $i$th agent is able to approach the finishing line $\mathcal{C}\left( {{\mathbf{p}}_{\text{f}}}\right) $ is given. As the function $V$ is not a Lyapunov function, here we use the \textit{invariant set theorem} \cite[p. 69]{Slotine(1991)} to do the analysis.
\begin{itemize}[leftmargin=*]
	\item 
	Firstly, we will study the property of the function $V$. Let $%
	\Omega=\left
	\{\mathbf{p}_{1},\cdots, \mathbf{p}_{M}:{V}\leq l\right \} ,$ $l>0. $ As there exists $V_{\text{m},ij}, V_{\text{tl},i},V_{\text{tr},i}>0$, ${V}\leq l$ implies $\sum_{i=1}^{M}V_{\text{l},i}\leq l.$ Furthermore, according to \textit{%
		Lemma 1(iii)}, $\Omega$ is bounded. When $\left \Vert \left[\mathbf{p}_{1},\cdots, \mathbf{p}_{M}\right]\right \Vert \rightarrow \infty,$ then $\sum_{i=1}^{M}V_{\text{l},i}\rightarrow \infty$ according to \textit{Lemma 1(ii)},
	namely ${V}\rightarrow \infty$. Therefore the function $V$ satisfies the
	condition that the invariant set theorem requires.
	
	Secondly, we will find the largest invariant set and show that all agents can pass $\mathcal{C}\left( {{\mathbf{p}}_{\text{f}}}\right) $. It is obtained that ${\dot{V}}={0}$ if and only if 
	\begin{align*}
		\mathbf{P}_{\text{t}}\text{sat}\left(k_1\tilde{\mathbf{p}}_{\text{l},i},v_{\text{m},i}\right) -\sum_{j=1,i \neq j}^{M}b_{ij}\tilde{\mathbf{{p}}}_{\text{m},ij}+\left(\frac{\partial V_{\text{tl},i}}{\partial \mathbf{p}_i }\right)^{\text{T}} \notag \\
		+\left(\frac{\partial V_{\text{tr},i}}{\partial \mathbf{p}_i }\right)^{\text{T}}=\mathbf{0}, 
	\end{align*}
	where $i=1,\cdots ,M$. Then in this case, we have $\mathbf{v}_{\text{c},i}=\mathbf{0}$ according
	to ({\ref{controller1}}). Consequently, the system cannot get 
	``stuck'' at an equilibrium point other than $\mathbf{v}_{\text{c},i}=\mathbf{0}$. 
	
	\item
	Finally, we will prove that no agent will get ``stuck''. Let the $1$st agent be ahead of the swarm, namely it is the closest to $\mathcal{C}\left( {{\mathbf{p}}_{\text{f}}}\right)$. When there exists $\mathbf{v}_{\text{c},1}=\mathbf{0}$, we examine the following equation related to the 1st agent that 
	\begin{equation}
		k_1{\kappa}_{v_{\text{m},1}}\mathbf{P}_\text{t}\mathbf{\tilde{p}}_\text{l,1}-\sum_{j=2}^{M}{b}_{1j}{\mathbf{\tilde{p}}}_{\text{m,}1j}+\left(\frac{\partial V_{\text{tl},1}}{\partial \mathbf{p}_1 }\right)^{\text{T}}+\left(\frac{\partial V_{\text{tr},1}}{\partial \mathbf{p}_1 }\right)^{\text{T}}=\mathbf{0}.
		\label{equilibriumTh5_1st}
	\end{equation}
	Since the $1$st agent is ahead, we have 
	\begin{equation}
		-\mathbf{t}_{\text{c}}^{\text{T}}\mathbf{\tilde{p}}_{\text{m,}1j}\leq 0,
		\label{1st}
	\end{equation}
	where ``='' holds if and only if the $j$th agent is as ahead as the $1$st agent. Then, multiplying the term $-\mathbf{t}_{\text{c}}^{\text{T}}$ on the left side of (\ref{equilibriumTh5_1st}) results in
	\begin{align*}
		-{{k}_{1}{\kappa }_{v_{\text{m},1}}}\mathbf{t}_{\text{c}}^{\text{T}}{\mathbf{%
				P}_{\text{t}}{\mathbf{\tilde{p}}}_{\text{l,}1}^{\prime}}=&-\mathbf{t}_{\text{c}}^{%
			\text{T}}\underset{j=2}{\overset{M}{\sum }}b_{1j}\mathbf{\tilde{p}}_{\text{m,
			}1j}+\mathbf{t}_{\text{c}}^{\text{T}}\left(\frac{\partial V_{\text{tl},i}}{\partial \mathbf{p}_i }\right)^{\text{T}}\\
		&+\mathbf{t}_{\text{c}}^{\text{T}}\left(\frac{\partial V_{\text{tl},i}}{\partial \mathbf{p}_i }\right)^{\text{T}}.
	\end{align*}
	Since  (\ref{DirL2}), (\ref{DirR2}), (\ref{1st}) hold for the 1st agent, we have $-\mathbf{t}_{\text{c}}^{\text{T}}{\mathbf{P}_{\text{t}}{\mathbf{\tilde{p}}}_{\text{l,}1}}={-\mathbf{t}_{\text{c}}^{\text{T}}{{\mathbf{\tilde{p}}}_{\text{l,}1}}\leq 0}$.
	As we have  $-\mathbf{t}_{\text{c}}^{\text{T}}{{\mathbf{\tilde{p}}}_{\text{l,}1}}%
	\left( 0\right) >{0}$ in the beginning according to \textit{Assumption 1}, owing to the
	continuity, given ${\epsilon }_{\text{0}}>0,$ there must exist a time $%
	t_{11}>0$ such that $-\mathbf{t}_{\text{c}}^{\text{T}}\mathbf{\tilde{p}}_{\text{l,}1}\left(t\right)\leq {\epsilon }_{\text{0}}$ at $t\geq t_{11}.$ At the time $t_{11},$ the $1$st agent is removed from the trapezoid virtual tube according to \textit{Assumption 2}. The problem left is to consider the $M-1$ agents, namely $2$nd, $3$rd,
	..., $M$th agents. We can repeat the analysis above to conclude this proof. $\square $  
\end{itemize}

\bibliographystyle{IEEEtran}
\bibliography{tube}

\end{document}